\documentclass[sigconf,screen]{acmart}

\usepackage{algorithm}
\usepackage{algpseudocode}
\usepackage{graphicx}
\usepackage{siunitx}
\usepackage{booktabs}    
\usepackage{multirow}
\usepackage{siunitx}
\usepackage{dblfloatfix}
\AtBeginDocument{%
  }

\copyrightyear{2025}
\acmYear{2025}
\setcopyright{acmlicensed}\acmConference[MMSports '25]{Proceedings of the 8th International ACM Workshop on Multimedia Content Analysis in Sports}{October 27--28, 2025}{Dublin, Ireland}
\acmBooktitle{Proceedings of the 8th International ACM Workshop on Multimedia Content Analysis in Sports (MMSports '25), October 27--28, 2025, Dublin, Ireland}
\acmDOI{10.1145/3728423.3759400} \acmISBN{979-8-4007-1835-9/2025/10}

\begin{document}

\title[A Dataset and Algorithm for Basketball Tracking with Identification and Pose Estimation]{TrackID3x3: A Dataset and Algorithm for Multi-Player Tracking with Identification and Pose Estimation in 3x3 Basketball Full-court Videos}

\author{Kazuhiro Yamada}
\authornote{The three authors contributed equally to this research.}
\author{Li Yin}
\authornotemark[1]
\author{Qingrui Hu}
\authornotemark[1]
\affiliation{%
  \institution{Nagoya University}
  \city{Nagoya}
  \country{Japan}
}

\author{Ning Ding}
\affiliation{%
  \institution{Nagoya Institute of Technology}
  \city{Nagoya}
  \country{Japan}}

\author{Shunsuke Iwashita}
\affiliation{%
  \institution{Nagoya University}
  \city{Nagoya}
  \country{Japan}
}

\author{Jun Ichikawa}
\affiliation{%
 \institution{Shizuoka University}
 \city{Shizuoka}
 \country{Japan}}

\author{Kiwamu Kotani}
\affiliation{%
  \institution{Ryutsu Keizai University}
  \city{Ryugasaki}
  \country{Japan}}

\author{Calvin Yeung}
\author{Keisuke Fujii}
\affiliation{%
  \institution{Nagoya University}
  \city{Nagoya}
  \country{Japan}}

\renewcommand{\shortauthors}{Yamada et al.}

\begin{abstract}
  Multi-object tracking, player identification, and pose estimation are fundamental components of sports analytics, essential for analyzing player movements, performance, and tactical strategies. However, existing datasets and methodologies primarily target mainstream team sports such as soccer and conventional 5-on-5 basketball, often overlooking scenarios involving fixed-camera setups commonly used at amateur levels, less mainstream sports, or datasets that explicitly incorporate pose annotations. 
  In this paper, we propose the TrackID3x3 dataset, the first publicly available comprehensive dataset specifically designed for multi-player tracking, player identification, and pose estimation in 3x3 basketball scenarios. The dataset comprises three distinct subsets (Indoor fixed-camera, Outdoor fixed-camera, and Drone camera footage), capturing diverse full-court camera perspectives and environments. We also introduce the Track-ID task, a simplified variant of the game state reconstruction task that excludes field detection and focuses exclusively on fixed-camera scenarios. To evaluate performance, we propose a baseline algorithm called Track-ID algorithm, tailored to assess tracking and identification quality. Furthermore, our benchmark experiments, utilizing recent multi-object tracking algorithms (e.g., CAMELTrack) and top-down pose estimation methods (HRNet, RTMPose, and SwinPose), demonstrate robust results and highlight remaining challenges. Our dataset and evaluation benchmarks provide a solid foundation for advancing automated analytics in 3x3 basketball.
Dataset and code will be available at \url{https://github.com/open-starlab/TrackID3x3}. 
\end{abstract}

\ccsdesc[500]{Computing methodologies~Computer vision tasks~Activity recognition and understanding}
\ccsdesc[300]{Computer systems organization~Architectures~Other architectures~Special purpose systems}

\keywords{Dataset, Sports, Multi-object tracking, Human pose estimation}

\maketitle

\begin{figure}[!t]
\centering
\includegraphics[width=0.48\textwidth]{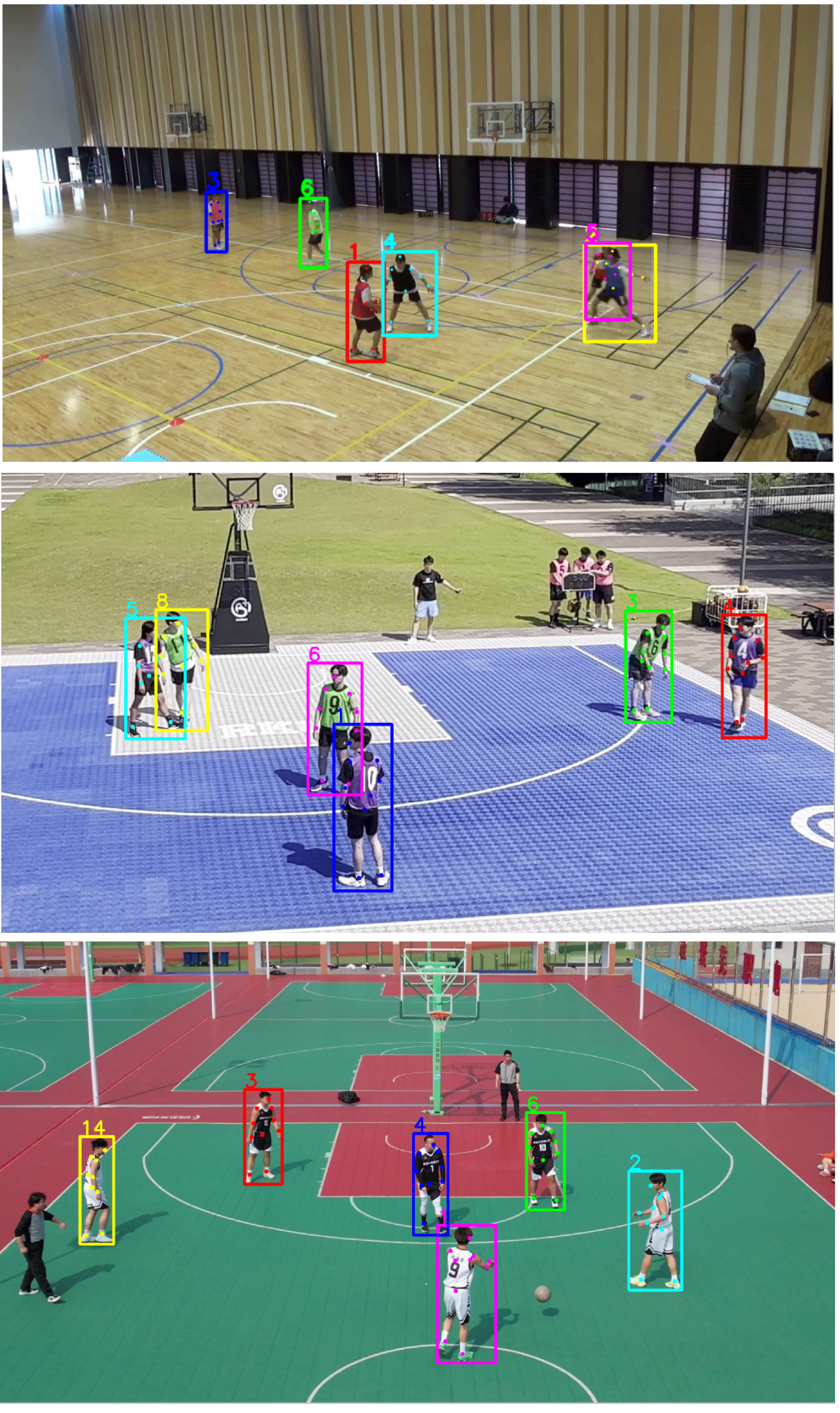}
\vspace{-25pt}
\caption{Examples of TrackID3x3 dataset with bounding boxes and pose (10 keypoints). We provide indoor fixed-camera (top: Indoor), outdoor fixed-camera (middle: Outdoor), outdoor drone-camera (bottom: Drone) videos in 3x3 basketball.
We estimate the player coordinates on the court in the Indoor and Outdoor datasets.}
\label{fig:overview}
\end{figure}

\begin{table*}[!t]\centering
\begin{tabular}{lrrrrc}\toprule
Dataset & Frames & Sports & Pose & Place & Camera \\\midrule
SN-Tracking~\cite{cioppa2022soccernet} &225,375 &Soccer & $\times$ & outdoor & broadcast camera\\
DeepSport-Basketball \cite{ghasemzadeh2021deepsportlab} & 672 & Basketball & $\checkmark$ & indoor & broadcast camera
\\
DanceTrack~\cite{sun2022dancetrack} &105,855  &Dance & $\times$ & indoor/outdoor & various cameras 
\\
SportsMOT~\cite{cui2023sportsmot} &150,379  & Soccer/Basketball/Volleyball & $\times$ & indoor/outdoor & broadcast camera\\
SoccerTrack\cite{scott2022soccertrack} & 82,800  & Soccer & $\times$ & outdoor & fisheye/drone\\
TeamTrack \cite{scott2024teamtrack} & 279,900 & Soccer/Basketball/Handball & $\times$ & indoor/outdoor &  fisheye/drone \\
SN-GSR \cite{somers2024soccernet} & 150,000 & Soccer & $\times$ & outdoor & broadcast camera \\ 
\midrule
\textbf{TrackID3x3 (ours)} & 277,926 & 3x3 Basketball & $\checkmark$ & indoor/outdoor & smartphone/consumer/drone\\ 
\bottomrule
\end{tabular}
\caption{Comparative overview of MOT and related datasets in sports.}\label{tab:prior}
\vspace{-15pt}
\end{table*}

\vspace{-5pt}
\section{Introduction}
\label{sec:intro}
\vspace{0pt}
Recent advances in computer vision have developed sports analytics by enabling the automated extraction of critical information from video data, reducing manual intervention and enhancing analytical precision. Researchers have developed techniques for field registration (e.g., \cite{theiner2023tvcalib,gutierrez2024no}), multi-object tracking (MOT, e.g., \cite{yin2024enhanced,hu2024basketball,sun2024gta}), re-identification (Re-ID, e.g., \cite{mansourian2023multi,somers2023body,somers2024keypoint}), and pose estimation (e.g., \cite{xiong2022swin,xu2022vitpose,cheng2020higherhrnet}). These approaches also address common sports-specific challenges, such as frequent occlusions and the inherent similarity in visual features among players, which complicate accurate data extraction. These will also support applications ranging from automated decision making and scoring to comprehensive performance analytics by converting raw broadcast footage into structured data.  However, the current research is predominantly focused on sports such as soccer, 5-on-5 basketball, and volleyball, while many sports—including 3x3 basketball—lack dedicated datasets and tailored methodologies.

Beyond soccer, the availability of tracking data enough for machine learning analytics in other team sports remains critically limited (note that, the datasets in sports computer vision areas are usually not sufficient for data analytics using machine learning). For instance, the only large-scale tracking dataset available for 5-on-5 basketball is derived from the 2015–16 NBA season, including 630 games; however, it has been outdated now and hinder further research progress. In contrast, soccer game state reconstruction \cite{somers2024soccernet}—which integrates field detection, player tracking, and identification in broadcast videos—has been pioneering and remains essential for subsequent analyses such as event prediction \cite{Decroos19,toda2022evaluation,yeung2025transformer,yeung2025openstarlab}, trajectory prediction \cite{lindstrom2020predicting,fujii2024decentralized}, and reinforcement learning \cite{nakahara2023action,fujii2023adaptive}. 
To enable such analytical techniques across sports, it is necessary to develop methods that can effectively process video footage captured by simple, fixed cameras (e.g., smartphones). Establishing robust image processing approaches for these conditions is expected to democratize automated analyses based on machine learning, including prediction with mathematical modeling \cite{Spearman18,kono2024mathematical,iwashita2024space}, evaluation \cite{ding2023estimation,fujii2024estimating}, and suggestion of optimal actions \cite{yeung2024strategic,nakahara2023action}. In particular, 3x3 basketball offers a promising research opportunity: its reduced number of players simplifies occlusion problems and streamlines problem formulation in computer vision. However, a significant challenge remains in the absence of annotated datasets for player positions and pose estimation in this sport.

In this paper, we propose TrackID3x3 dataset and algorithm for multi-player tracking with identification and pose estimation in 3x3 basketball full-court videos (see Figure \ref{fig:overview} and Table \ref{tab:prior}).
Our dataset, the first publicly available resource specifically for 3x3 basketball, includes subsets captured from Indoor (fixed camera), Outdoor (fixed camera), and Drone (drone camera) scenarios, providing diversity in terms of camera perspectives and playing environments. We also propose a new task termed Track-ID task, a simplified version of the game state reconstruction (GSR) task \cite{somers2024soccernet}, excluding field detection to target fixed-camera scenarios, thereby enabling straightforward processes (see Section \ref{sec:track-id_task}). Additionally, we establish baseline algorithms using a combination of recent tracking and identification methods. Our benchmark evaluation incorporates recent algorithms for MOT and 2D pose estimation.

The contributions of this paper are as follows.
(1) We propose a new comprehensive annotation dataset in 3x3 basketball full-court videos captured under diverse conditions—including indoor and outdoor with smartphone, drone, and consumer camera recordings—addressing the current lack of publicly available resources for this sport (Table \ref{tab:prior}).
(2) We examined the feasibility of fully automating multi-player tracking with identification—a simplified variant of game state reconstruction \cite{somers2024soccernet}—in 3x3 basketball, a sport that offers more favorable conditions compared to soccer.
Our experimental results demonstrate the effectiveness of our baseline approach, achieving 85.53\% in the Indoor dataset and 71.03\% in the Outdoor dataset for the proposed TI-HOTA metric (described in Section \ref{sec:algorithm}). 
(3) We present and validate them as the pose dataset, an essential component for future subsequent sports analytics. We establish benchmarks for pose estimation on our dataset by evaluating a range of models from conventional methods to recent top-down transformer-based models, reporting over 77\% in all datasets
in terms of mean Percentage of Detected Joints (PDJ).

\begin{table*}[!ht]\centering
\begin{tabular}{lrrrrrc}\toprule
Subset in TrackID3x3 & Frames   & BBoxes    & Pose       & Place & Camera & Resolution \\\midrule
Indoor dataset              & 7,531    & 45,186     & 2,601    & indoor   & Consumer-grade (Sony HDR-CX680) & 1280 × 720 \\
Outdoor dataset             & 143,276   &  859,656   & 3,600   & outdoor    & Smartphone (iPhone 13) &  3840 × 2160  \\
Drone dataset              &  127,119        &  762,714         & 500    & outdoor    & Drone (DJI Air 2S) & 3840 × 2160 \\
\bottomrule
\end{tabular}
\caption{Overview of our TrackID3x3 Dataset. The unit of pose is frames and that of resolutions is pixel.}\label{tab:ours}
\vspace{-15pt}
\end{table*}

\vspace{-5pt}
\section{Related Work}
\label{sec:related}
\vspace{-3pt}
\noindent \textbf{Multi-player tracking in sports.}
In recent years, significant progress has been made in building datasets. Early efforts provided soccer player location data using multi-view camera \cite{d2009semi}, while subsequent datasets combined high-resolution panorama videos with local positioning system data \cite{pettersen2014soccer}. Publicly available large-scale broadcast video datasets such as SoccerNet \cite{cioppa2022soccernet,somers2024soccernet} and SoccerDB \cite{jiang2020soccerdb} have further propelled research in player tracking. In basketball, datasets including APIDIS and SPIROUDOME \cite{de2008distributed, lu2017light} have been introduced, alongside other sports datasets for handball \cite{biermann2021unified} and volleyball \cite{msibrahiCVPR16deepactivity}. Recent datasets such as SoccerNet-Tracking \cite{cioppa2022soccernet} and SportsMOT \cite{cui2023sportsmot} utilize unedited broadcast footage, while SoccerTrack \cite{scott2022soccertrack} and its extension TeamTrack \cite{scott2024teamtrack} offer full-pitch multi-sport tracking. Furthermore, virtual environments like Google Research Football \cite{kurach2020google} and Soccersynth \cite{qin2025soccersynth} provide synthetic data that facilitate algorithm benchmarking. 

In recent MOT, tracking‐by‐detection is a popular approach, where high-quality object detection forms the backbone for subsequent data association and tracking. The first step is the accurate identification of players and the ball in each video frame. Early methods relied on background subtraction, multi-scale feature detection, and edge detection techniques \cite{mackowiak2010football, cheshire2015player, direkoglu2018player}. More recently, deep learning–based detectors such as RetinaNet \cite{lin2017focal}, CenterNet \cite{duan2019centernet}, and various versions of YOLO \cite{redmon2016you} have significantly improved detection accuracy. The second step associates detections across consecutive frames, which is challenging in sports due to frequent occlusions, rapid motion, and visual similarity among players. Motion-based methods, such as SORT \cite{bewley2016simple} and its extensions like BoT-SORT \cite{aharon2022bot}, employ Kalman filters and data association techniques to predict object trajectories. Deep appearance-based methods, exemplified by DeepSORT \cite{wojke2017simple} and techniques leveraging Deep-EIoU \cite{huang2024iterative}, use convolutional neural networks to extract robust visual features that help maintain consistent identities despite occlusion and similar uniforms. In addition, the recently proposed CAMELTrack \cite{somers2025CAMELTrack} incorporates an association method based on a learnable module rather than the heuristic associations like the above methods. 
Recent works in soccer \cite{maglo2022efficient,majeed2024mv}, basketball \cite{hu2024basketball,yin2024enhanced}, hockey \cite{vats2023player}, and multi-sports \cite{huang2023observation,sun2024gta,Khanna2025SportMamba,Stanczyk2025NoTrainYetGain}, combine these strategies and their own unique methods to tackle the unique challenges of sports tracking, including complex movements, similar appearances, and severe occlusions. The proposed dataset provides experimental settings that simplify issues in basketball player tracking, such as multi-player occlusion \cite{hu2024basketball}, as well as camera positions and bibs that are not found in existing benchmark datasets.

\noindent \textbf{Player identification in sports.} 
Player identification, including Re-ID (the process of consistently matching and maintaining a player's identity across different frames and camera views) is fundamental to sports analytics. In non-team sports, marathon, ultra running, and running datasets \cite{ben2012racing,penate2020tgc20reid,suzuki2024runner} provide images or videos of athletes annotated with or without bib numbers. In team sports, re-ID datasets have been proposed in basketball \cite{van2022deepsportradarv1}, ice hockey \cite{koshkina2024general, yingnan2020MHPTD}, and soccer  \cite{cioppa2022scaling}. Recently, the SoccerNet GSR dataset \cite{somers2024soccernet} provided a comprehensive dataset integrating field registration, tracking, and player identification. 

Player Re-ID methods have evolved from relying solely on traditional visual cues such as jersey color and number \cite{ivankovic2014automatic-basket-color}. Early image-level methods focused on facial and local appearance features, while more recent work employs convolutional neural networks and part-based methods \cite{akan2023reidentifying}. Advanced techniques now incorporate contrastive learning and transformer-based models \cite{Koshkina2021-pn-hockey-classification, Vats2022-lq-hockey-tfidentification}, as well as multi-task frameworks for joint Re-ID, team affiliation, and role classification (PRTreID) \cite{mansourian2023multi}. In GSR \cite{somers2024soccernet} baseline, player identification was performed with PRTreID \cite{mansourian2023multi} to produce team and role-aware ReID embeddings, and MMOCR \cite{kuang2021mmocr} for jersey number recognition. We use a general framework \citep{koshkina2024general} for jersey number recognition and color histograms, and relative position on the court to determine team affiliation. 

\noindent \textbf{Pose estimation in sports.} 
Pose information is fundamental in sports analytics, offering detailed insights into players’ movements for precise performance assessment. In MOT, pose annotations provide fine-grained structural cues for Re-ID or tracklet association. In scenarios with occlusions or similar uniforms, keypoint-based features help reduce ID switches. Many studies have leveraged pose information to improve MOT \cite{iqbal2017posetrack, yin2024enhanced}. While our pose annotations are not solely for training MOT baselines, they offer multimodal supervision (bounding box + pose) to support future research and pose-assisted MOT approaches.

In sports contexts, the LSP dataset contains 1,000 pose-labeled images \cite{johnson2010clustered}, the Sport Image dataset has 1,300 \cite{wang2011learning}, and COCO-WholeBody provides comprehensive whole-body annotations \cite{jin2020whole}. Domain-specific datasets include 3DSP \cite{yeung2024autosoccerpose} and DeepSportLab for basketball, which offers ball detection, player segmentation, and pose estimation (672 frames) \cite{ghasemzadeh2021deepsportlab}. Our pose annotations aim to provide sufficient samples for method development and validation, covering diverse scenes, viewpoints, and occlusions. Based on our experience, the 6,701 annotated frames are sufficient for research and evaluation (Table \ref{tab:ours}).

Recent advances in 2D pose estimation have been primarily driven by top-down methods in sports (e.g., \cite{yeung2024autosoccerpose, yin2024enhanced}), where the process begins with detecting individual persons followed by keypoint localization. The top-down paradigm consistently yields higher precision in keypoint estimation in sports scenes characterized by frequent occlusions and rapid motion. DeepPose \cite{Toshev2014DeepPose} pioneered this approach, and subsequent work such as HRNet \cite{Sun2019Deep} further improved accuracy by preserving high-resolution representations throughout the network. More recent top-down systems, including RTMPose \cite{jiang2023rtmpose} and SwinPose \cite{xiong2022swin}, leverage transformer-based architectures to further refine keypoint predictions. Although bottom-up methods like DeepCut \cite{pishchulin2016deepcut} and OpenPose \cite{cao2018openpose} as well as emerging end-to-end approaches \cite{shi2022end, liu2023group} provide efficient alternatives, we focus our evaluation solely on top-down methods because bounding boxes are given in our datasets.

\vspace{-5pt}
\section{TrackID3x3 Dataset}
\label{sec:dataset}
\vspace{-3pt}

As described in Table \ref{tab:ours}, we propose TrackID3x3 dataset composed of three subsets: Indoor, Outdoor, and Drone datasets (the novelty of our dataset is described in Introduction and Table \ref{tab:prior}.
In this section, we describe the data collection and annotations in this order. 

\subsection{Dataset collection}
\vspace{-3pt}
\noindent \textbf{Indoor dataset.}
As described in \cite{ichikawa2024analysis}, Indoor dataset was constructed a fixed-camera 3x3 basketball videos consisting of 42 videos totaling 7,531 frames for six players on the court. The six female university basketball players regularly participate in official 5-on-5 games. 
The measurement was conducted in a university gymnasium (half-court of 5-on-5 basketball court) using a Sony HDR-CX680 camera ($1280\times720$). Thus, only this subset has lower resolution compared to other subsets, with a lot of motion blur, making it relatively difficult to detect players in some scenes.
For the mini-game design, offensive and defensive teams were fixed with players being assigned to specific roles. The ball handler of \#1 is fixed, and the player begins without directly passing to \#3. The start positions of the two offensive players (\#1 and \#2) are called the two-guard position used in a 5-on-5 game. Additionally, the start position of each offensive player is assigned to maintain a certain distance for observing the coordination process that the role of intervention decision and adjustment involves in the other players. On the defensive team, \#1 and \#2 confront offensive \#1 and \#2, and all three defenders’ starting positions are voluntary. The above information will help with accurate player identification. 
Written informed consent was obtained from all the participants according to \cite{ichikawa2024analysis}. 

\noindent \textbf{Outdoor dataset.}
Outdoor dataset was constructed as a fixed-camera 3x3 basketball videos consisting of 12 videos totaling 143,276 frames for six players on the court. Fourteen male university basketball players, who regularly participate in official 5-on-5 games, created four teams (three players for each team; the remaining two players were reserved for accidents). 
A total of 12 games were played following a double round-robin format (i.e., each team participated in 6 games), and the four teams wore bibs in distinct colors: green, purple, pink, and yellow.
The measurement was conducted in a university outdoor 3x3 court using two smartphone cameras (iPhone 13: $3840\times2160$); but in this dataset we used only one camera. 
The 3x3 game was played according to the official 3x3 rules\footnote{FIBA 3x3 official rules \url{https://fiba3x3.com/en/rules.html}\label{fn:3x3rules}} except that the game was limited to 5 minutes without timeouts and without a knockout rule (21 points).
This measurement was approved by the ethical committee of Nagoya University and Ryutsu Keizai University 
and written informed consent was obtained from all the participants.

\noindent \textbf{Drone dataset.}
Drone dataset was constructed from a drone camera 3x3 basketball videos consisting of 6 videos (92 min. in total) but excluding out-of-play sections, 127,119 frames were extracted focusing on the six players on the court.
Sixteen male university basketball players, who regularly participate in official 5-on-5 games, created four teams (three players for each team; the remaining four players were reserved for accidents). 
A total of 6 games were played following a round-robin format (i.e., each team participated in 3 games). Since only black, brown and white colors were available, one team wore black or brown jersey while the other team wore white jersey to easily distinguish between the teams.
The measurement was conducted in a university outdoor 3x3 court using two drone cameras (DJI Air 2S: $3840\times2160$), but in this dataset, we used only one drone camera. There are some camera movements, but it is not as dramatic as in broadcast footage in existing benchmark datasets, and the players are in the camera's field of view in almost every frame.
The 3x3 game was played according to the official 3x3 rules\footref{fn:3x3rules}.
This measurement was approved by the ethical committee of Anhui Normal University, China, 
and written informed consent was obtained from all the participants.

\vspace{-5pt}
\subsection{Annotation}
\vspace{-3pt}
\noindent \textbf{Field detection.}
In the Indoor dataset, field information was provided by \cite{ichikawa2024analysis}, which was manually annotated based on the basketball court keypoints.
Annotated keypoints in the center of a 5-centimeter wide court line, the court size was $9.50 \times 15.05$ in the Indoor dataset and $11.05 \times 15.05$ in the Outdoor dataset (depth $\times$ width [m], width corresponds to the length of the end line). 
Similarly, we manually annotated the field information in Outdoor dataset. 
However, in Drone dataset, although the field of view was generally stable, slight movements occurred due to wind and other environmental influences, and thus ground truth field information was not annotated. 
While field detection is a crucial component of the overall GSR task \cite{somers2024soccernet} (in particular, significantly impacting the holistic metric called GS-HOTA), this study focuses solely on the Track-ID task using fixed-camera data. 
Consequently, only the Indoor and Outdoor datasets include field information, whereas the Drone dataset is used exclusively for tracking.

\noindent \textbf{Bounding boxes.}
In the Indoor dataset, bounding boxes (bboxes) were provided by \cite{ichikawa2024analysis}. 
In the Outdoor dataset and Drone dataset, approximately half of the data was annotated manually, and the remaining half was annotated using a method adopted in a previous study \cite{xu2024finesports}, in which the output of the tracking algorithm was manually corrected (BoT-SORT-ReID \cite{aharon2022bot} was used as the tracking algorithm in this study).
For all datasets, unique IDs for each player were assigned to enable tracking and identification over time.

\noindent \textbf{Body keypoints.}
In a subset of frames from three datasets, we manually annotated 10 human body keypoints per player. These keypoints include the head, shoulders, elbows, wrists, ankles, and the center, defined as the midpoint of the hips. Compared to commonly used 2D human pose estimation datasets that include 17 keypoints, we reduced annotation complexity and eliminated unnecessary keypoints by omitting the eyes and ears. Additionally, we simplified the lower body representation by retaining only the ankles and the hip midpoint.

\begin{figure*}[!t]
\centering
\includegraphics[width=1\textwidth]{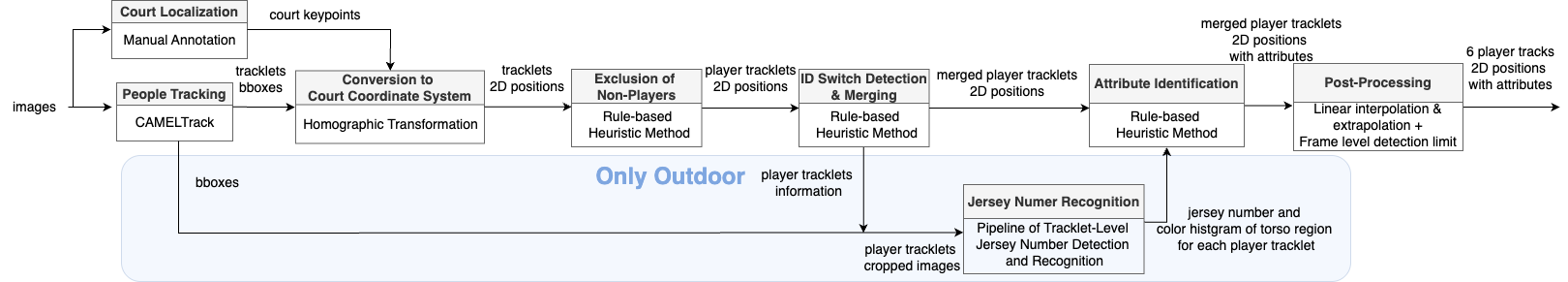}
\vspace{-10pt}
\caption{Our Track-ID baseline algorithm. Note that the court localization is performed manually and others can be automatically done.}
\label{fig:baseline}
\vspace{-10pt}
\end{figure*}

\vspace{-5pt}
\section{TrackID3x3 Baseline Algorithms}
\label{sec:algorithm}
\vspace{-3pt}
Here, we introduce the TrackID3x3 baseline algorithms employed to evaluate the performance of tracking, identification, and pose estimation methods on our dataset.
First, we introduce a new Track-ID task, and its baseline algorithm, as well as tracking (Appendix \ref{app:MOTdrone}) and 2D pose estimation baselines. 
\vspace{-5pt}
\subsection{Track-ID Task}
\label{sec:track-id_task}
\vspace{-3pt}
One of our experimental tasks is a Track-ID task to obtain the location and identification information of players on the court from fixed-camera videos. It is similar to the recent GSR task \citep{somers2024soccernet}, but the most significant difference is that the camera is fixed and shows the entire court. Our task is relatively easy and accurate to carry out court localization by manually annotating keypoints on a single image in the video. Another important difference is that referees are not the object of this task; in 3x3 basketball, since the two referees only move back and forth in their respective areas, their movements have little effect on the state of the game, and there is less incentive to track them than in soccer. Since only on-court players are targeted, the maximum number of detections required in each frame is determined to be 6. 

Note that there is a crucial difference between the Indoor and Outdoor datasets, in terms of player identification: the presence or absence of players' bib numbers. This is because the numbers were not printed on the bibs of the Indoor dataset, and this difference also accounts for the difference in sub-tasks performed between the two datasets. In addition, for the Outdoor dataset, the player is uniquely determined by two pieces of information: team affiliation and jersey number, because players on the same team wear the same color bibs with different numbers. Hence, the task formulation is also different in the two datasets. 

We formalize the Track-ID task in 3x3 basketball according to the GSR task as follows. For each on-court player detection $d_i^t$ ($i \in \{1, 2, ..., 6\}$) at frame $t$ in a video with a total number of frames $T$, predict its location and identification information. The information to be predicted in the Indoor dataset is,
\begin{equation}
\label{Indoor_d_i^t}
    d_i^t=\{\underbrace{\text{\textit{court\_x, court\_y}}}_{\text{localization}},\;
    \underbrace{\text{\textit{team, initial\_position}}}_{\text{identification}}\},
\end{equation}
where \textit{court\_x} and \textit{court\_y} are the $x, y$ coordinates in the court coordinate system, $\textit{team}$ $\in\{\text{offense, defense}\}$ indicates whether the player was on the offensive or defensive team at the start of the video, and $\textit{initial\_position}$ $\in \{\text{top, left, right}\}$ indicates the position at the start of the video. For each attribute of the ground truth $\textit{initial\_position}$, the top is given to two players who start the offense and defend her by the top of the key (no check-ball is played and the former holds the ball from the beginning of the video). The left and right are given relative to the coordinates of the other players. The information that should be predicted in the Outdoor dataset is,
\begin{equation}
\label{Outdoor_d_i^t}
    d_i^t=\{\underbrace{\text{\textit{court\_x, court\_y}}}_{\text{localization}},\;
    \underbrace{\text{\textit{team, jersey\_number}}}_{\text{identification}}\},
\end{equation}
where \textit{jersey\_number} is the jersey number of the bib, otherwise the same as the Indoor dataset. With the above two definitions, both Indoor and Outdoor on-court players can be uniquely identified.

\noindent \textbf{Track-ID Metric.}
The Track-ID task is evaluated by a metric called Track-ID HOTA (TI-HOTA). This is essentially equivalent to GS-HOTA but is given a different name to clarify the difference in the tasks performed. GS-HOTA is a metric that improves the similarity score for matching predictions and ground-truth from Higher Order Tracking Accuracy (HOTA) \cite{luiten2021hota} to evaluate GSR, calculated only for pairs of predictions and ground-truth that are misaligned by less than the distance tolerance parameter $\tau \,[m]$ and have completely matching identification. Although the formulas are exactly the same, the difference between TI-HOTA and GS-HOTA is in the subtasks: manual or automatic court localization and the different identification information required. 

\noindent Therefore, TI-HOTA is calculated as follows:
\begin{equation}
\text{TI-HOTA} = \frac{1}{19} \sum_{\alpha \in \{0.05,0.10,\dots,0.95\}} \sqrt{\operatorname{TI-DetA}_{\alpha} \times \operatorname{TI-AssA}_{\alpha}},\\
\end{equation}
where 
\begin{gather}
\operatorname{TI-DetA}_{\alpha} = \frac{|TP_\alpha|}{|TP_\alpha| + |FP_\alpha| + |FN_\alpha|}, \\
\operatorname{TI-AssA}_{\alpha} = \frac{1}{|TP_\alpha|} \sum_{c \in \{{TP_\alpha}\}}\mathcal{A}_\alpha(c), \\
\mathcal{A}_\alpha(c) = \frac{|TPA_\alpha(c)|}{|TPA_\alpha(c)| + |FPA_\alpha(c)| + |FNA_\alpha(c)|}.
\
\end{gather}
$|TP_\alpha|, |FP_\alpha|$, and $|FN_\alpha|$ are the number of true positives, false positives, and false negatives at threshold $\alpha$ of similarity score as described below, and $|TPA_\alpha(c)|, |FPA_\alpha(c)|$ and $|FNA_\alpha(c)|$ are the number of true positive associations, false positive associations, and false negative associations. Note that $c$ is a true positive pair that matched at least once in all frames based on the similarity score: 
\begin{gather}
\operatorname{Sim}_{\mathrm{TI\text{-}HOTA}}(P,G) = \operatorname{LocSim}(P,G) \times \operatorname{IdSim}(P,G), \\
\operatorname{LocSim}(P, G)=e^{\ln (0.05) \frac{\|P-G\|_2^2}{\tau^2}}, \\
\operatorname{IdSim}(P, G)= \begin{cases}1 & \text { if all attributes match, } \\
0 & \text { otherwise. }\end{cases}
\end{gather}
where $P$, $G$ are prediction and ground truth.

In addition to the TI-HOTA, this paper reports its sub-scores ($\operatorname{TI-DetA}$ and $\operatorname{TI-AssA}$) for detail evalutation.

\begin{table*}[t]
\centering
\scalebox{0.98}{
\begin{minipage}{0.48\linewidth}
    \centering
    \scalebox{0.83}{
    \begin{tabular}{ccccc}
      \toprule
      \textbf{Dataset} & \textbf{Length [s]} & \textbf{TI-HOTA (\%)↑} & \textbf{TI-DetA (\%)↑} & \textbf{TI-AssA (\%)↑} \\ 
      \midrule
      Indoor   & 8.98 $\pm$ 3.24  & 85.53 $\pm$ 7.09  & 84.86 $\pm$ 8.01  & 86.23 $\pm$ 6.15 \\
      Outdoor  & 40.11 $\pm$ 34.23 & 71.03 $\pm$ 17.79 & 67.15 $\pm$ 20.12 & 75.51 $\pm$ 15.65 \\
      \bottomrule
    \end{tabular}
    }
  \end{minipage}%
   \hspace{0.03\textwidth}
   \begin{minipage}{0.48\linewidth}
    \centering
    \scalebox{0.83}{
    \begin{tabular}{ccccc}
      \toprule
      \textbf{Dataset} & \textbf{Length [s]} & \textbf{TI-HOTA (\%)↑} & \textbf{TI-DetA (\%)↑} & \textbf{TI-AssA (\%)↑} \\ 
      \midrule
      Indoor   & 8.98 $\pm$ 3.24   & 84.45 $\pm$ 7.44  & 83.60 $\pm$ 8.42  & 85.33 $\pm$ 6.44 \\
      Outdoor  & 40.11 $\pm$ 34.23 & 66.20 $\pm$ 18.07 & 61.97 $\pm$ 19.86 & 71.08 $\pm$ 16.42 \\
      \bottomrule
    \end{tabular}
    }
  \end{minipage}
}

\caption{Experimental results of the Track-ID task; Left: w/ ID Switch Merging, Right: w/o ID Switch Merging. Note that the mean ± SD at $\tau=1$ is presented.}
\label{tab:Track-ID_results}
\vspace{-5pt}
\end{table*}

\vspace{-5pt}
\subsection{Track-ID Baseline Algorithm}
\vspace{-3pt}
We propose a baseline to execute this task and verify its performance. Figure \ref{fig:baseline} shows an overview of the baseline algorithm. A  detailed description is provided as follows.
Note that the court localization is manually performed and described in Section \ref{sec:dataset}. 

\noindent \textbf{People Detection and Tracking.}
We use CAMELTrack \citep{somers2025CAMELTrack} with a pre-trained YOLOX \citep{yolox2021} detecter without fine-tuning to track people in the videos. This tracking algorithm is one of the few methods that has reported a HOTA score of over 80\% on the SportsMOT test set \cite{cui2023sportsmot}, and was selected for its high accuracy and code availability. The weights of the association module were pre-trained using SportsMOT \cite{cui2023sportsmot}. The weights used in YOLOX are the same as those used during annotation, but the post-processing settings were changed to reduce detection power. This is partly because it is the same detector used for annotation, so as to avoid using exactly the same detection results, but also because in basketball player tracking, it may be effective to use a detecter that does not have overly strong detection power (the detail explanation is described in Section \ref{sec:conclusion}).

\noindent \textbf{Conversion to Court Coordinate System.}
To convert the bboxes of the persons to court coordinates, homographic transformation was performed using the annotated court keypoints. The position coordinate of each tracklet is calculated using the midpoint of the bottom edge of the bboxes. 

\noindent \textbf{Exclusion of Non-Players.}
Since the images include people who are playing and those who are not, it is necessary to distinguish them. We employ a rule-based heuristic method based on court position coordinates. Exclusion criteria are divided into detection level and tracklet level. There is one exclusion criterion for detection level: detections more than 3 meter away from the outer lines of the court are excluded. On the other hand, tracklet level exclusion is performed based on three criteria: tracks that do not exist continuously for 10 or more frames on the court, tracks that exist in more than half of the frames in the area along the end line, from 3 meters outside the court to 1 meter inside the court, and tracks that exist in more than half of the frames outside the extension lines of the two long sides of the paint area and more than 10 meter away from the end line (the area near the so-called coffin corner). The latter two criteria are intended to exclude referees. For datasets with jersey numbers, referees can be excluded using the results of jersey number recognition described later, but to lighten the processing load as much as possible, we use simple rules to exclude them here. 

\noindent \textbf{ID Switch Detection \& Merging.}
After narrowing down the player's tracklets, apply the module that detects and merges the ID Switches. The purpose of this module is to merge fragmented tracklets for the same player. Specifically, a tracklet ID that appeared midway through a video clip is recognized as an ID switch and merged with a tracklet ID that existed from the beginning if the simultaneous existence of those IDs frame count is below the threshold (for details, refer to Algorithm \ref{pseudo code of IDs Det & Merg}). Although it cannot handle ID exchanges between players (described later) and requires all players to be constantly visible on camera, it has fewer parameters to adjust than Global Tracklet Association (GTA) \cite{sun2024gta}, which includes two clustering algorithms, and does not require the calculation of ReID features, resulting in lower computational costs. In this study, the only parameter to be adjusted, $T_{\text{overlap}}$, was set to 10 for Indoor and 50 for Outdoor, taking into account the average length of the video clips.

\begin{algorithm}
\small
\caption{ID Switch Detection \& Merging}
\label{pseudo code of IDs Det & Merg}
\begin{algorithmic}[1]
\Require DataFrame $D$ containing tracklets, overlap threshold $T_{\text{overlap}}$, total frame count of video clip $F_{\max}$
\State $t_{0} \gets \min(D.\text{frame\_id})$
\State $\text{origIDs} \gets \{\,i \mid D[\text{frame\_id} = t_{0}].\text{tracklet\_id} = i\}$
\State $\text{allIDs}  \gets \text{unique trackletIDs in }D$
\State $\text{newIDs}  \gets \text{allIDs} \setminus \text{origIDs}$
\ForAll{$\text{newID}$ in $\text{newIDs}$}
  \State $\text{candidates} \gets []$
  \State $\text{framesNew}  \gets \{\,f \mid D.\text{tracklet\_id}=\text{newID}  \land \text{on\_court}(f)\}$
  \ForAll{$\text{origID}$ in $\text{origIDs}$}
    \State $\text{framesOrig} \gets \{\,f \mid D.\text{tracklet\_id}=\text{origID} \land \text{on\_court}(f)\}$
    \State $\text{overlap} \gets |\text{framesOrig} \cap \text{framesNew}|$
    \If{$\text{overlap} \geq T_{\text{overlap}}$}
      \State \textbf{continue}
    \EndIf
    \State $\text{unionFrames} \gets \text{framesOrig} \cup \text{framesNew}$
    \State $\text{missing} \gets F_{\max} - |\text{unionFrames}|$
    \State $\text{cost} \gets \text{overlap} + \text{missing}$
    \State Append $(\text{origID}, \text{cost})$ to $\text{candidates}$
  \EndFor
  \If{$\text{candidates}$ is empty}
    \State \textbf{continue}
  \EndIf
  \State $\text{bestID} \gets$ origID in $\text{candidates}$ with minimum cost
  \State Relabel all rows where tracklet\_id = newID to tracklet\_id = bestID
  \State Drop duplicate rows on (frame\_id, tracklet\_id), keeping the first
\EndFor
\end{algorithmic}
\end{algorithm}

\noindent \textbf{Jersey Number Recognition.}
As mentioned earlier, there is a difference between Indoor and Outdoor in terms of the presence or absence of bibs numbers. Therefore, only Outdoor requires recognition of jersey numbers. We used a trucklet-level jersey number recognition pipeline \cite{koshkina2024general} to obtain one jersey number prediction result for each merged player tracklet. At the same time, we calculated the 3D color histogram for each tracklet using the cutout image of the torso region, which is an intermediate product. At the same time, we calculated the 3D color histogram for each tracklet using the cropped images of the torso region, which is an intermediate product. The color histogram has 8 × 8 × 8 bin and the median value over up to 100 frames from the smallest frame number.

\noindent \textbf{Attribute Identification.}
Not only does the Indoor dataset not have information on the player's jersey number, but all the players wear different colored bibs. Hence, the method of identifying the players' team affiliation is practically limited to determining it from her relative position at the beginning of the video. Specifically, in the opening frame of the video, three pairs were created by pairing players detected at close distances to each other, with the one closer to the midpoint of the end line as the defense and the one farther away as the offense. The top, left, and right were determined relative to the midpoint of each pair's position.

As for the Outdoor, it is more complicated: first, exclude tracklets whose jersey numbers could not be recognized to colpletely e referees. Next, the tracklet closest to a certain position in the first frame is the offense: the midpoint of the outer court line parallel to the end line in the case of videos where the offense starts at the top, or the midpoint of the free throw line in the case of videos where the offense begins with a free throw. Then, in the case of a video with the offense starting at the top, the tracklet closest to the determined offensive one is the defense, and then the two tracklets with the highest color histgram similarity based on Jensen-Shannon divergence to the offensive one are the offense and all the rest are the defense. In the case where the video begins with a free throw, the two tracklets with the most similar color histograms to the first determined offensive shall be the offense and all the remainder shall be the defensive. 

\noindent \textbf{Post-Processing.}
Finally, perform post-processing to produce the final output. Here, linear interpolation, end point extrapolation, and frame level detection number limitation are applied. Linear interpolation simply fills in the gaps in the tracklet linearly. Endpoint extrapolation fills in missing tracklets at the beginning or end of a video based on the difference between the two consecutive detections immediately after or before. The frame level detection limitation excludes detections in more than 6 detections, starting with those track with the fewest on-court frames, to ensure that there are 6 players or fewer.

\vspace{-5pt}
\subsection{Pose Estimation Baselines}
\vspace{-3pt}
In all datasets, we evaluate the performance of RTMPose \cite{jiang2023rtmpose}, HRNet \cite{Sun2019Deep}, and SwinPose \cite{xiong2022swin} to estimate the human pose. These models are common baseline models for top-down 2D pose estimation with diverse architectures.

We use The Percentage of Detected Joints (PDJ) \cite{Toshev2014DeepPose}, widely used for evaluating the accuracy of 2D human pose estimation. It measures the proportion of correctly detected keypoints based on a distance threshold relative to the torso diameter. This metric is useful for assessing pose estimation models with varying degrees of localization precision \cite{yeung2024autosoccerpose}.
In this evaluation metric, a keypoint is deemed correctly detected if the normalized distance between the predicted and ground truth keypoints is less than 0.5. Furthermore, PDJ allows for the evaluation of pose estimation accuracy under different levels of localization precision by adjusting the threshold. The area under the PDJ curve (AUC) is computed to assess performance across various threshold values.

For normalization, we designate the nose as the head reference point and define the center point as the midpoint between the left and right hips. The torso length is determined as the Euclidean distance between the midpoint of the left and right shoulders and the center.

\vspace{-8pt}
\section{Experiments}
\label{sec:experiments}
\vspace{-3pt}
Here, we show the experimental results of our baseline algorithms on our datasets. 

\vspace{-8pt}
\subsection{Performance in Track-ID task}
\vspace{-3pt}
Experiments were conducted on the Track-ID task on fixed camera datasets, Indoor and Outdoor, as described above. We experimented with this task mainly on the in-play portion: Indoor was all frames of each video, while Outdoor excluded the out-of-play portion and divided the video of one game into several segments. Two types of video segments exist: one starting with a check-ball at the top and the other starting with a free throw (the former number of videos is 74, and the latter number of videos is 22). The numbers of video frames in Indoor and Outdoor datasets 179.38 $\pm$ 64.77 and 1202.84 $\pm$ 1025.92 (mean $\pm$ standard deviation), respectively. 
The evaluation metrics are TI-HOTA and its sub-scores TI-DetA, TI-AssA as described above. For the distance tolerance parameter, $\tau$ of TI-HOTA, $\tau = 1$ was adopted given the ratio to the football pitch. 

Tables \ref{tab:Track-ID_results} show the results, including the mean and standard deviation, when ID Switch Merging was performed and when it was not performed, demonstrating the effectiveness and robustness of the corresponding module. Note that ID Switch based on two identification information is not reported because it did not occur in the proposed baseline .

\begin{figure}[h]
    \centering
    \includegraphics[width=0.7\linewidth]{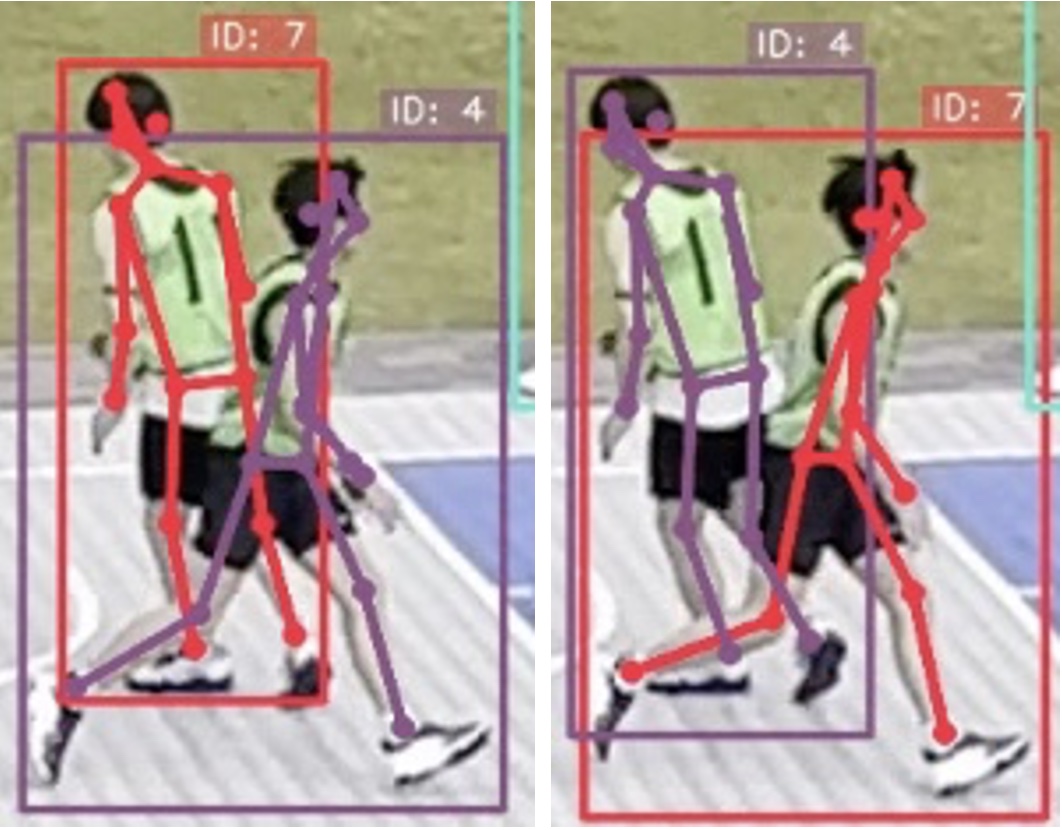}
    \vspace{-12pt}
    \caption{Exmaple of ID exchange during tracking in Outdoor dataset. It occurs during occlusion and is a significant contributing factor to the decrease in the TI‑HOTA score.}
    \vspace{-10pt}
    \label{fig:ID_exchange}
\end{figure}

It is difficult to compare the Indoor and Outdoor results simply , but the better performance of Indoor can be explained to some extent by the shorter video length and the easier identification of attributes. This is because if ID exchange between players occurs in tracking (see Figure \ref{fig:ID_exchange}), the attributes will no longer be correct from then on, and the match between the prediction and the ground truth will fail for a large part of the video clip.

\begin{table*}[htbp] \centering
\scalebox{1}[1]{
\begin{tabular}{ll S[table-format=3.2] S[table-format=3.2] S[table-format=3.2] 
                    S[table-format=3.2] S[table-format=3.2] S[table-format=3.2] 
                    S[table-format=3.2] S[table-format=3.2]}
\toprule
Dataset & Model & {Head} & {Sho} & {EL} & {Wri} & {Cen}  & {Ank}  & {PDJ} & {AUC} \\
\midrule
\multirow{3}{*}{Indoor}
& RTMPose \cite{jiang2023rtmpose} &68.69\% & \bfseries 87.26\% & \bfseries 71.99\% & \bfseries 70.02\% & \bfseries 77.87\% & \bfseries 78.29\% &\bfseries 75.69\% & \bfseries 37.84\% \\
& HRNet \cite{Sun2019Deep}     & \bfseries 69.51\% & 86.76\% & 68.62\% & 66.36\% & 77.31\% & 75.60\% & \ 74.03\% & 37.01\% \\
& SwinPose\cite{xiong2022swin} & 69.20\% & 86.13\% & 64.19\% & 67.48\% & 77.27\% & 74.44\% & 73.12\% & 36.56\% \\
\midrule
\multirow{3}{*}{Outdoor} 
& RTMPose \cite{jiang2023rtmpose} & \bfseries 88.82\% & 93.57\% & \bfseries 91.03\% & \bfseries 85.86\% & 89.40\% & 87.60\% & \bfseries 89.43\% & 45.12\% \\
& HRNet \cite{Sun2019Deep}    & 88.39\% & \bfseries 93.88\% & 89.53\% & 85.65\% & \bfseries 91.43\% & 89.09\% & 88.51\% & \bfseries 45.21\% \\
& SwinPose\cite{xiong2022swin} & 88.21\% & 93.14\% & 89.19\% & 85.20\% & 90.16\% & \bfseries 90.16\% & 89.27\% & 45.03\% \\
\midrule
\multirow{3}{*}{Drone}
& RTMPose \cite{jiang2023rtmpose} & 85.57\% & 85.60\% & \bfseries 85.17\% & \bfseries 80.64\% & 77.59\% & 84.30\% & 83.45\% & 42.07\% \\
& HRNet \cite{Sun2019Deep}      & \bfseries 85.99\% & 86.09\% & 84.53\% & 79.69\% & \bfseries 84.60\% & \bfseries 84.75\% & \bfseries 84.07\% & \bfseries 42.38\% \\
& SwinPose\cite{xiong2022swin} & 85.49\% & \bfseries 86.30\% & 84.43\% & 79.72\% & 83.05\% & 84.70\% & 83.88\% & 42.28\% \\
\bottomrule
\end{tabular}
}
\caption{2D pose estimation models performance on all datasets, ranked by PDJ. The best-performing result for each dataset is highlighted in bold.
Sho, El, Wri, Cen, and Ank represent the average PDJ of Shoulder, Elbow, Wrist, Center, and Ankle, respectively. The Center is defined as the midpoint between the left and right hips. The PDJ column represents the mean PDJ across all 10 keypoints. AUC represents the area under the PDJ curve as the threshold varies from 0 to 0.5 in increments of 0.01, with a maximum value of 0.5.}
\label{tab:Pos_eva}
\vspace{-15pt}
\end{table*}

\vspace{-3pt}
\subsection{2D Pose Estimation Performance}
\vspace{-2pt}
We also evaluated the performance of 2D pose estimation models in our all datasets. We evaluated three models without fine-tuning as baselines. Since each player in our datasets is annotated with only 10 keypoints, while the baseline models output 17 keypoints in COCO \cite{lin2015microsoft} keypoints format, we select the corresponding 10 keypoints for evaluation to ensure consistency. 

The evaluation results are summarized in Table~\ref{tab:Pos_eva}.
All models achieved their highest PDJ scores on the Outdoor dataset, but performance declined on the Drone dataset, with the Indoor dataset exhibiting the lowest accuracy. The lower performance observed in the Indoor dataset can be primarily attributed to its lower image resolution (1280 × 720) compared to the Outdoor and Drone datasets (3820 × 2160), which may result in blurred keypoints and reduced model accuracy.

\begin{figure}[h]
    \centering
    \includegraphics[width=0.8\linewidth]{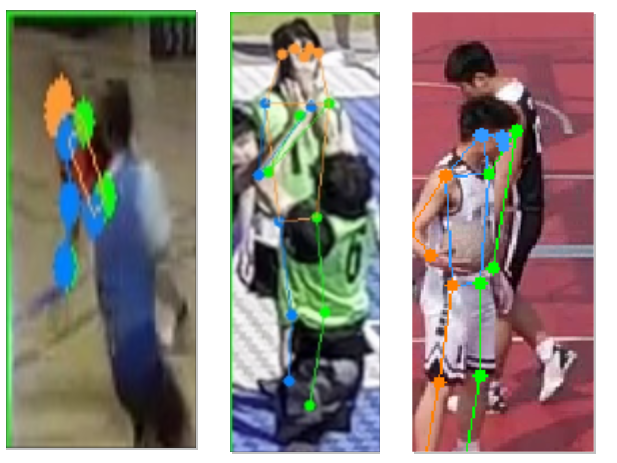}
    \vspace{-15pt}
    \caption{Examples of players with low PDJ across three datasets. From left to right, the images correspond to the Indoor, Outdoor, and Drone datasets. The players positioned at the back are the primary targets for keypoint detection. However, due to occlusion, keypoint prediction for the obstructed player results in partial or complete failure.}
    \vspace{-10pt}
    \label{fig:pose result}
\end{figure}

A comparison between different keypoints across datasets, ankles, shoulders, and the center exhibit the most stable and high detection accuracy in all scenarios. In contrast, the elbows and wrists in the upper limbs demonstrate weaker performance, probably due to occlusion and motion blur.
In our TrackID3x3 dataset, RTMPose\cite{jiang2023rtmpose} achieved the best performance in the Outdoor dataset, with an average PDJ of 89.43\%. This result is highly comparable to the average PDJ of 89.51\% reported for RTMPose in AutoSoccerPose \cite{yeung2024autosoccerpose}.
For illustration, Figure \ref{fig:pose result} presents examples of players with lower prediction accuracy in the three datasets.

\vspace{-8pt}
\section{Conclusion}
\label{sec:conclusion}
\vspace{-3pt}
In this paper, we proposed the TrackID3x3 dataset and its baseline algorithms in 3x3 basketball. Our benchmark experiments demonstrated robust results and highlight remaining challenges. Our dataset and evaluation benchmarks provide a solid foundation for advancing automated analytics in 3x3 basketball. 

Although we were unable to provide quantitative or qualitative evidence, tracking methods based on Tracking by Detection, such as CAMELTrack, may actually increase mis-tracking during occlusion when using detectors with excessively high detection power. This is because the bounding boxes of hidden players may be nearly identical to those of players in the foreground (i.e., there are many pairs with very high IoU), which can easily lead to incorrect associations. This is particularly pronounced in sports like basketball, where occlusions occur frequently, and could be a contributing factor to ID exchange, one of the most critical challenges in the Track-ID task. The proposed baseline takes this trade-off into account and deliberately uses a detector with reduced detection power (though still sufficiently effective for tracking). However, ID exchange still occurred, so to truly address this issue, it may be necessary to adopt tracking algorithms that include ID Switch countermeasures during occlusion (such as OccluTrack \cite{Gao2023OccluTrack}) or introduce post-processing methods that include tracklet splitting (such as GTA \cite{sun2024gta}). The latter currently uses only appearance features for split, so there may be cases where the split does not work properly when ID exchange occurs between teammates (because teammates often have similar appearances) but by utilizing prediction results of jersey number, it may be possible to achieve more effective splitting and merging, potentially bringing us closer to resolving this issue.

\noindent{\bfseries Acknowledgements.}
This work was financially supported by JST PRESTO Grant Number JPMJPR20CA, JSPS KAKENHI Grant Number 23H03282, JP24K23889 and Mizuno Sports Promotion Foundation.

\bibliographystyle{ACM-Reference-Format}
\bibliography{acmart-primary/acmart}


\begin{thebibliography}{89}


\ifx \showCODEN    \undefined \def \showCODEN     #1{\unskip}     \fi
\ifx \showISBNx    \undefined \def \showISBNx     #1{\unskip}     \fi
\ifx \showISBNxiii \undefined \def \showISBNxiii  #1{\unskip}     \fi
\ifx \showISSN     \undefined \def \showISSN      #1{\unskip}     \fi
\ifx \showLCCN     \undefined \def \showLCCN      #1{\unskip}     \fi
\ifx \shownote     \undefined \def \shownote      #1{#1}          \fi
\ifx \showarticletitle \undefined \def \showarticletitle #1{#1}   \fi
\ifx \showURL      \undefined \def \showURL       {\relax}        \fi
\providecommand\bibfield[2]{#2}
\providecommand\bibinfo[2]{#2}
\providecommand\natexlab[1]{#1}
\providecommand\showeprint[2][]{arXiv:#2}

\bibitem[Aharon et~al\mbox{.}(2022)]%
        {aharon2022bot}
\bibfield{author}{\bibinfo{person}{Nir Aharon}, \bibinfo{person}{Roy Orfaig}, {and} \bibinfo{person}{Ben-Zion Bobrovsky}.} \bibinfo{year}{2022}\natexlab{}.
\newblock \showarticletitle{BoT-SORT: Robust associations multi-pedestrian tracking}.
\newblock \bibinfo{journal}{\emph{arXiv preprint arXiv:2206.14651}} (\bibinfo{year}{2022}).
\newblock


\bibitem[Akan and Varl{\i}(2023)]%
        {akan2023reidentifying}
\bibfield{author}{\bibinfo{person}{Sara Akan} {and} \bibinfo{person}{Song{\"u}L Varl{\i}}.} \bibinfo{year}{2023}\natexlab{}.
\newblock \showarticletitle{Reidentifying soccer players in broadcast videos using Body Feature Alignment Based on Pose}. In \bibinfo{booktitle}{\emph{Proceedings of the 2023 4th International Conference on Computing, Networks and Internet of Things}}. \bibinfo{pages}{440--444}.
\newblock


\bibitem[Ben-Ami et~al\mbox{.}(2012)]%
        {ben2012racing}
\bibfield{author}{\bibinfo{person}{Idan Ben-Ami}, \bibinfo{person}{Tali Basha}, {and} \bibinfo{person}{Shai Avidan}.} \bibinfo{year}{2012}\natexlab{}.
\newblock \showarticletitle{Racing Bib Numbers Recognition.}. In \bibinfo{booktitle}{\emph{Proceedings of the British Machine Vision Conference}}. \bibinfo{pages}{1--10}.
\newblock


\bibitem[Bewley et~al\mbox{.}(2016)]%
        {bewley2016simple}
\bibfield{author}{\bibinfo{person}{Alex Bewley}, \bibinfo{person}{Zongyuan Ge}, \bibinfo{person}{Lionel Ott}, \bibinfo{person}{Fabio Ramos}, {and} \bibinfo{person}{Ben Upcroft}.} \bibinfo{year}{2016}\natexlab{}.
\newblock \showarticletitle{Simple online and realtime tracking}. In \bibinfo{booktitle}{\emph{2016 IEEE International Conference on Image Processing (ICIP)}}. IEEE, \bibinfo{pages}{3464--3468}.
\newblock


\bibitem[Biermann et~al\mbox{.}(2021)]%
        {biermann2021unified}
\bibfield{author}{\bibinfo{person}{Henrik Biermann}, \bibinfo{person}{Jonas Theiner}, \bibinfo{person}{Manuel Bassek}, \bibinfo{person}{Dominik Raabe}, \bibinfo{person}{Daniel Memmert}, {and} \bibinfo{person}{Ralph Ewerth}.} \bibinfo{year}{2021}\natexlab{}.
\newblock \showarticletitle{A unified taxonomy and multimodal dataset for events in invasion games}. In \bibinfo{booktitle}{\emph{Proceedings of the 4th International Workshop on Multimedia Content Analysis in Sports}}. \bibinfo{pages}{1--10}.
\newblock


\bibitem[Cao et~al\mbox{.}(2018)]%
        {cao2018openpose}
\bibfield{author}{\bibinfo{person}{Zhe Cao}, \bibinfo{person}{Gines Hidalgo}, \bibinfo{person}{Tomas Simon}, \bibinfo{person}{Shih-En Wei}, {and} \bibinfo{person}{Yaser Sheikh}.} \bibinfo{year}{2018}\natexlab{}.
\newblock \showarticletitle{Open{P}ose: realtime multi-person 2{D} pose estimation using {P}art {A}ffinity {F}ields}. In \bibinfo{booktitle}{\emph{arXiv preprint arXiv:1812.08008}}.
\newblock


\bibitem[Cheng et~al\mbox{.}(2020)]%
        {cheng2020higherhrnet}
\bibfield{author}{\bibinfo{person}{Bowen Cheng}, \bibinfo{person}{Bin Xiao}, \bibinfo{person}{Jingdong Wang}, \bibinfo{person}{Honghui Shi}, \bibinfo{person}{Thomas~S Huang}, {and} \bibinfo{person}{Lei Zhang}.} \bibinfo{year}{2020}\natexlab{}.
\newblock \showarticletitle{Higherhrnet: Scale-aware representation learning for bottom-up human pose estimation}. In \bibinfo{booktitle}{\emph{Proceedings of the IEEE/CVF Conference on Computer Vision and Pattern Recognition}}. \bibinfo{pages}{5386--5395}.
\newblock


\bibitem[Cheshire et~al\mbox{.}(2015)]%
        {cheshire2015player}
\bibfield{author}{\bibinfo{person}{Evan Cheshire}, \bibinfo{person}{Min-Chun Hu}, {and} \bibinfo{person}{Ming-Hsiu Chang}.} \bibinfo{year}{2015}\natexlab{}.
\newblock \showarticletitle{Player tracking and analysis of basketball plays}. In \bibinfo{booktitle}{\emph{European Conference of Computer Vision}}.
\newblock


\bibitem[Cioppa et~al\mbox{.}(2022a)]%
        {cioppa2022scaling}
\bibfield{author}{\bibinfo{person}{Anthony Cioppa}, \bibinfo{person}{Adrien Deliege}, \bibinfo{person}{Silvio Giancola}, \bibinfo{person}{Bernard Ghanem}, {and} \bibinfo{person}{Marc Van~Droogenbroeck}.} \bibinfo{year}{2022}\natexlab{a}.
\newblock \showarticletitle{Scaling up SoccerNet with multi-view spatial localization and re-identification}.
\newblock \bibinfo{journal}{\emph{Scientific Data}} \bibinfo{volume}{9}, \bibinfo{number}{1} (\bibinfo{year}{2022}), \bibinfo{pages}{355}.
\newblock


\bibitem[Cioppa et~al\mbox{.}(2022b)]%
        {cioppa2022soccernet}
\bibfield{author}{\bibinfo{person}{Anthony Cioppa}, \bibinfo{person}{Silvio Giancola}, \bibinfo{person}{Adrien Deliege}, \bibinfo{person}{Le Kang}, \bibinfo{person}{Xin Zhou}, \bibinfo{person}{Zhiyu Cheng}, \bibinfo{person}{Bernard Ghanem}, {and} \bibinfo{person}{Marc Van~Droogenbroeck}.} \bibinfo{year}{2022}\natexlab{b}.
\newblock \showarticletitle{Soccernet-tracking: Multiple object tracking dataset and benchmark in soccer videos}. In \bibinfo{booktitle}{\emph{Proceedings of the IEEE/CVF Conference on Computer Vision and Pattern Recognition}}. \bibinfo{pages}{3491--3502}.
\newblock


\bibitem[Cui et~al\mbox{.}(2023)]%
        {cui2023sportsmot}
\bibfield{author}{\bibinfo{person}{Yutao Cui}, \bibinfo{person}{Chenkai Zeng}, \bibinfo{person}{Xiaoyu Zhao}, \bibinfo{person}{Yichun Yang}, \bibinfo{person}{Gangshan Wu}, {and} \bibinfo{person}{Limin Wang}.} \bibinfo{year}{2023}\natexlab{}.
\newblock \showarticletitle{Sports{MOT}: A large multi-object tracking dataset in multiple sports scenes}. In \bibinfo{booktitle}{\emph{Proceedings of the IEEE/CVF International Conference on Computer Vision}}. \bibinfo{pages}{9921--9931}.
\newblock


\bibitem[De~Vleeschouwer et~al\mbox{.}(2008)]%
        {de2008distributed}
\bibfield{author}{\bibinfo{person}{Christophe De~Vleeschouwer}, \bibinfo{person}{Fan Chen}, \bibinfo{person}{Damien Delannay}, \bibinfo{person}{Christophe Parisot}, \bibinfo{person}{Christophe Chaudy}, \bibinfo{person}{Eric Martrou}, \bibinfo{person}{Andrea Cavallaro}, {et~al\mbox{.}}} \bibinfo{year}{2008}\natexlab{}.
\newblock \showarticletitle{Distributed video acquisition and annotation for sport-event summarization}.
\newblock \bibinfo{journal}{\emph{New European Media Summit}}  \bibinfo{volume}{8} (\bibinfo{year}{2008}).
\newblock


\bibitem[Decroos et~al\mbox{.}(2019)]%
        {Decroos19}
\bibfield{author}{\bibinfo{person}{Tom Decroos}, \bibinfo{person}{Lotte Bransen}, \bibinfo{person}{Jan Van~Haaren}, {and} \bibinfo{person}{Jesse Davis}.} \bibinfo{year}{2019}\natexlab{}.
\newblock \showarticletitle{Actions speak louder than goals: Valuing player actions in soccer}. In \bibinfo{booktitle}{\emph{Proceedings of the 25th ACM SIGKDD Conference on Knowledge Discovery and Data Mining}}. \bibinfo{pages}{1851--1861}.
\newblock


\bibitem[Ding et~al\mbox{.}(2024)]%
        {ding2023estimation}
\bibfield{author}{\bibinfo{person}{Ning Ding}, \bibinfo{person}{Kazuya Takeda}, \bibinfo{person}{Wenhui Jin}, \bibinfo{person}{Yingjiu Bei}, {and} \bibinfo{person}{Keisuke Fujii}.} \bibinfo{year}{2024}\natexlab{}.
\newblock \showarticletitle{Estimation of control area in badminton doubles with pose information from top and back view drone videos}.
\newblock \bibinfo{journal}{\emph{Multimedia Tools and Applications}} \bibinfo{volume}{83}, \bibinfo{number}{8} (\bibinfo{year}{2024}), \bibinfo{pages}{24777--24793}.
\newblock


\bibitem[Direkoglu et~al\mbox{.}(2018)]%
        {direkoglu2018player}
\bibfield{author}{\bibinfo{person}{Cem Direkoglu}, \bibinfo{person}{Melike Sah}, {and} \bibinfo{person}{Noel O'connor}.} \bibinfo{year}{2018}\natexlab{}.
\newblock \showarticletitle{Player detection in field sports}.
\newblock \bibinfo{journal}{\emph{Machine Vision and Applications}} \bibinfo{volume}{29}, \bibinfo{number}{2} (\bibinfo{year}{2018}), \bibinfo{pages}{187--206}.
\newblock


\bibitem[D'Orazio et~al\mbox{.}(2009)]%
        {d2009semi}
\bibfield{author}{\bibinfo{person}{Tiziana D'Orazio}, \bibinfo{person}{Marco Leo}, \bibinfo{person}{Nicola Mosca}, \bibinfo{person}{Paolo Spagnolo}, {and} \bibinfo{person}{Pier~Luigi Mazzeo}.} \bibinfo{year}{2009}\natexlab{}.
\newblock \showarticletitle{A semi-automatic system for ground truth generation of soccer video sequences}. In \bibinfo{booktitle}{\emph{2009 Sixth IEEE International Conference on Advanced Video and Signal Based Surveillance}}. IEEE, \bibinfo{pages}{559--564}.
\newblock


\bibitem[Duan et~al\mbox{.}(2019)]%
        {duan2019centernet}
\bibfield{author}{\bibinfo{person}{Kaiwen Duan}, \bibinfo{person}{Song Bai}, \bibinfo{person}{Lingxi Xie}, \bibinfo{person}{Honggang Qi}, \bibinfo{person}{Qingming Huang}, {and} \bibinfo{person}{Qi Tian}.} \bibinfo{year}{2019}\natexlab{}.
\newblock \showarticletitle{Centernet: Keypoint triplets for object detection}. In \bibinfo{booktitle}{\emph{Proceedings of the IEEE/CVF International Conference on Computer Vision}}. \bibinfo{pages}{6569--6578}.
\newblock


\bibitem[Fujii et~al\mbox{.}(2024a)]%
        {fujii2024decentralized}
\bibfield{author}{\bibinfo{person}{Keisuke Fujii}, \bibinfo{person}{Naoya Takeishi}, \bibinfo{person}{Yoshinobu Kawahara}, {and} \bibinfo{person}{Kazuya Takeda}.} \bibinfo{year}{2024}\natexlab{a}.
\newblock \showarticletitle{Decentralized policy learning with partial observation and mechanical constraints for multiperson modeling}.
\newblock \bibinfo{journal}{\emph{Neural Networks}}  \bibinfo{volume}{171} (\bibinfo{year}{2024}), \bibinfo{pages}{40--52}.
\newblock


\bibitem[Fujii et~al\mbox{.}(2024b)]%
        {fujii2024estimating}
\bibfield{author}{\bibinfo{person}{Keisuke Fujii}, \bibinfo{person}{Koh Takeuchi}, \bibinfo{person}{Atsushi Kuribayashi}, \bibinfo{person}{Naoya Takeishi}, \bibinfo{person}{Yoshinobu Kawahara}, {and} \bibinfo{person}{Kazuya Takeda}.} \bibinfo{year}{2024}\natexlab{b}.
\newblock \showarticletitle{Estimating counterfactual treatment outcomes over time in complex multi-agent scenarios}.
\newblock \bibinfo{journal}{\emph{IEEE Transactions on Neural Networks and Learning Systems}} (\bibinfo{year}{2024}), \bibinfo{pages}{1--15}.
\newblock


\bibitem[Fujii et~al\mbox{.}(2023)]%
        {fujii2023adaptive}
\bibfield{author}{\bibinfo{person}{Keisuke Fujii}, \bibinfo{person}{Kazushi Tsutsui}, \bibinfo{person}{Atom Scott}, \bibinfo{person}{Hiroshi Nakahara}, \bibinfo{person}{Naoya Takeishi}, {and} \bibinfo{person}{Yoshinobu Kawahara}.} \bibinfo{year}{2023}\natexlab{}.
\newblock \showarticletitle{Adaptive action supervision in reinforcement learning from real-world multi-agent demonstrations}.
\newblock \bibinfo{journal}{\emph{arXiv preprint arXiv:2305.13030}} (\bibinfo{year}{2023}).
\newblock


\bibitem[Gao et~al\mbox{.}(2023)]%
        {Gao2023OccluTrack}
\bibfield{author}{\bibinfo{person}{Jianjun Gao}, \bibinfo{person}{Yi Wang}, \bibinfo{person}{Kim-Hui Yap}, \bibinfo{person}{Kratika Garg}, {and} \bibinfo{person}{Boon~Siew Han}.} \bibinfo{year}{2023}\natexlab{}.
\newblock \bibinfo{title}{OccluTrack: Rethinking Awareness of Occlusion for Enhancing Multiple Pedestrian Tracking}.
\newblock
\showeprint{arXiv:2309.10360}


\bibitem[Ge et~al\mbox{.}(2021)]%
        {yolox2021}
\bibfield{author}{\bibinfo{person}{Zheng Ge}, \bibinfo{person}{Songtao Liu}, \bibinfo{person}{Feng Wang}, \bibinfo{person}{Zeming Li}, {and} \bibinfo{person}{Jian Sun}.} \bibinfo{year}{2021}\natexlab{}.
\newblock \showarticletitle{YOLOX: Exceeding YOLO Series in 2021}.
\newblock \bibinfo{journal}{\emph{arXiv preprint arXiv:2107.08430}} (\bibinfo{year}{2021}).
\newblock


\bibitem[Ghasemzadeh et~al\mbox{.}(2021)]%
        {ghasemzadeh2021deepsportlab}
\bibfield{author}{\bibinfo{person}{Seyed~Abolfazl Ghasemzadeh}, \bibinfo{person}{Gabriel Van~Zandycke}, \bibinfo{person}{Maxime Istasse}, \bibinfo{person}{Niels Sayez}, \bibinfo{person}{Amirafshar Moshtaghpour}, {and} \bibinfo{person}{Christophe De~Vleeschouwer}.} \bibinfo{year}{2021}\natexlab{}.
\newblock \showarticletitle{DeepSportLab: a Unified Framework for Ball Detection, Player Instance Segmentation and Pose Estimation in Team Sports Scenes}. In \bibinfo{booktitle}{\emph{The 32nd British Machine Vision Conference}}.
\newblock


\bibitem[Guti{\'e}rrez-P{\'e}rez and Agudo(2024)]%
        {gutierrez2024no}
\bibfield{author}{\bibinfo{person}{Marc Guti{\'e}rrez-P{\'e}rez} {and} \bibinfo{person}{Antonio Agudo}.} \bibinfo{year}{2024}\natexlab{}.
\newblock \showarticletitle{No Bells Just Whistles: Sports Field Registration by Leveraging Geometric Properties}. In \bibinfo{booktitle}{\emph{Proceedings of the IEEE/CVF Conference on Computer Vision and Pattern Recognition}}. \bibinfo{pages}{3325--3334}.
\newblock


\bibitem[Hu et~al\mbox{.}(2024)]%
        {hu2024basketball}
\bibfield{author}{\bibinfo{person}{Qingrui Hu}, \bibinfo{person}{Atom Scott}, \bibinfo{person}{Calvin Yeung}, {and} \bibinfo{person}{Keisuke Fujii}.} \bibinfo{year}{2024}\natexlab{}.
\newblock \showarticletitle{Basketball-SORT: An Association Method for Complex Multi-object Occlusion Problems in Basketball Multi-object Tracking}.
\newblock \bibinfo{journal}{\emph{Multimedia Tools and Applications}} (\bibinfo{year}{2024}).
\newblock


\bibitem[Huang et~al\mbox{.}(2023)]%
        {huang2023observation}
\bibfield{author}{\bibinfo{person}{Hsiang-Wei Huang}, \bibinfo{person}{Cheng-Yen Yang}, \bibinfo{person}{Samartha Ramkumar}, \bibinfo{person}{Chung-I Huang}, \bibinfo{person}{Jenq-Neng Hwang}, \bibinfo{person}{Pyong-Kun Kim}, \bibinfo{person}{Kyoungoh Lee}, {and} \bibinfo{person}{Kwangju Kim}.} \bibinfo{year}{2023}\natexlab{}.
\newblock \showarticletitle{Observation centric and central distance recovery for athlete tracking}. In \bibinfo{booktitle}{\emph{Proceedings of the IEEE/CVF Winter Conference on Applications of Computer Vision}}. \bibinfo{pages}{454--460}.
\newblock


\bibitem[Huang et~al\mbox{.}(2024)]%
        {huang2024iterative}
\bibfield{author}{\bibinfo{person}{Hsiang-Wei Huang}, \bibinfo{person}{Cheng-Yen Yang}, \bibinfo{person}{Jiacheng Sun}, \bibinfo{person}{Pyong-Kun Kim}, \bibinfo{person}{Kwang-Ju Kim}, \bibinfo{person}{Kyoungoh Lee}, \bibinfo{person}{Chung-I Huang}, {and} \bibinfo{person}{Jenq-Neng Hwang}.} \bibinfo{year}{2024}\natexlab{}.
\newblock \showarticletitle{Iterative scale-up expansioniou and deep features association for multi-object tracking in sports}. In \bibinfo{booktitle}{\emph{Proceedings of the IEEE/CVF Winter Conference on Applications of Computer Vision}}. \bibinfo{pages}{163--172}.
\newblock


\bibitem[Ibrahim et~al\mbox{.}(2016)]%
        {msibrahiCVPR16deepactivity}
\bibfield{author}{\bibinfo{person}{Mostafa~S. Ibrahim}, \bibinfo{person}{Srikanth Muralidharan}, \bibinfo{person}{Zhiwei Deng}, \bibinfo{person}{Arash Vahdat}, {and} \bibinfo{person}{Greg Mori}.} \bibinfo{year}{2016}\natexlab{}.
\newblock \showarticletitle{A Hierarchical Deep Temporal Model for Group Activity Recognition.}. In \bibinfo{booktitle}{\emph{2016 IEEE Conference on Computer Vision and Pattern Recognition (CVPR)}}.
\newblock


\bibitem[Ichikawa et~al\mbox{.}(2024)]%
        {ichikawa2024analysis}
\bibfield{author}{\bibinfo{person}{Jun Ichikawa}, \bibinfo{person}{Masatoshi Yamada}, {and} \bibinfo{person}{Keisuke Fujii}.} \bibinfo{year}{2024}\natexlab{}.
\newblock \showarticletitle{Analysis of coordinated group behavior based on role-sharing: Practical application from an experimental task to a 3-on-3 basketball game as a pilot study}.
\newblock \bibinfo{journal}{\emph{bioRxiv}} (\bibinfo{year}{2024}), \bibinfo{pages}{2024--09}.
\newblock


\bibitem[Iqbal et~al\mbox{.}(2017)]%
        {iqbal2017posetrack}
\bibfield{author}{\bibinfo{person}{Umar Iqbal}, \bibinfo{person}{Anton Milan}, {and} \bibinfo{person}{Juergen Gall}.} \bibinfo{year}{2017}\natexlab{}.
\newblock \showarticletitle{Posetrack: Joint multi-person pose estimation and tracking}. In \bibinfo{booktitle}{\emph{Proceedings of the IEEE Conference on Computer Vision and Pattern Recognition}}. \bibinfo{pages}{2011--2020}.
\newblock


\bibitem[Ivankovic et~al\mbox{.}(2014)]%
        {ivankovic2014automatic-basket-color}
\bibfield{author}{\bibinfo{person}{Zdravko Ivankovic}, \bibinfo{person}{Milos Rackovic}, {and} \bibinfo{person}{Miodrag Ivkovic}.} \bibinfo{year}{2014}\natexlab{}.
\newblock \showarticletitle{Automatic player position detection in basketball games}.
\newblock \bibinfo{journal}{\emph{Multimedia Tools and Applications}}  \bibinfo{volume}{72} (\bibinfo{year}{2014}), \bibinfo{pages}{2741--2767}.
\newblock


\bibitem[Iwashita et~al\mbox{.}(2024)]%
        {iwashita2024space}
\bibfield{author}{\bibinfo{person}{Shunsuke Iwashita}, \bibinfo{person}{Atom Scott}, \bibinfo{person}{Rikuhei Umemoto}, \bibinfo{person}{Ning Ding}, {and} \bibinfo{person}{Keisuke Fujii}.} \bibinfo{year}{2024}\natexlab{}.
\newblock \showarticletitle{Space evaluation based on pitch control using drone video in Ultimate}.
\newblock \bibinfo{journal}{\emph{arXiv preprint arXiv:}} (\bibinfo{year}{2024}).
\newblock


\bibitem[Jiang et~al\mbox{.}(2023)]%
        {jiang2023rtmpose}
\bibfield{author}{\bibinfo{person}{Tao Jiang}, \bibinfo{person}{Peng Lu}, \bibinfo{person}{Li Zhang}, \bibinfo{person}{Ningsheng Ma}, \bibinfo{person}{Rui Han}, \bibinfo{person}{Chengqi Lyu}, \bibinfo{person}{Yining Li}, {and} \bibinfo{person}{Kai Chen}.} \bibinfo{year}{2023}\natexlab{}.
\newblock \showarticletitle{Rtmpose: Real-time multi-person pose estimation based on mmpose}.
\newblock \bibinfo{journal}{\emph{arXiv preprint arXiv:2303.07399}} (\bibinfo{year}{2023}).
\newblock


\bibitem[Jiang et~al\mbox{.}(2020)]%
        {jiang2020soccerdb}
\bibfield{author}{\bibinfo{person}{Yudong Jiang}, \bibinfo{person}{Kaixu Cui}, \bibinfo{person}{Leilei Chen}, \bibinfo{person}{Canjin Wang}, {and} \bibinfo{person}{Changliang Xu}.} \bibinfo{year}{2020}\natexlab{}.
\newblock \showarticletitle{Soccerdb: A large-scale database for comprehensive video understanding}. In \bibinfo{booktitle}{\emph{Proceedings of the 3rd International Workshop on Multimedia Content Analysis in Sports}}. \bibinfo{pages}{1--8}.
\newblock


\bibitem[Jin et~al\mbox{.}(2020)]%
        {jin2020whole}
\bibfield{author}{\bibinfo{person}{Sheng Jin}, \bibinfo{person}{Lumin Xu}, \bibinfo{person}{Jin Xu}, \bibinfo{person}{Can Wang}, \bibinfo{person}{Wentao Liu}, \bibinfo{person}{Chen Qian}, \bibinfo{person}{Wanli Ouyang}, {and} \bibinfo{person}{Ping Luo}.} \bibinfo{year}{2020}\natexlab{}.
\newblock \showarticletitle{Whole-body human pose estimation in the wild}. In \bibinfo{booktitle}{\emph{Computer Vision--ECCV 2020: 16th European Conference, Glasgow, UK, August 23--28, 2020, Proceedings, Part IX 16}}. Springer, \bibinfo{pages}{196--214}.
\newblock


\bibitem[Johnson and Everingham(2010)]%
        {johnson2010clustered}
\bibfield{author}{\bibinfo{person}{Sam Johnson} {and} \bibinfo{person}{Mark Everingham}.} \bibinfo{year}{2010}\natexlab{}.
\newblock \showarticletitle{Clustered pose and nonlinear appearance models for human pose estimation}. In \bibinfo{booktitle}{\emph{Procedings of the British Machine Vision Conference}}. Aberystwyth, UK, \bibinfo{pages}{1--11}.
\newblock


\bibitem[Khanna et~al\mbox{.}(2025)]%
        {Khanna2025SportMamba}
\bibfield{author}{\bibinfo{person}{Dheeraj Khanna}, \bibinfo{person}{Jerrin Bright}, \bibinfo{person}{Yuhao Chen}, {and} \bibinfo{person}{John~S. Zelek}.} \bibinfo{year}{2025}\natexlab{}.
\newblock \bibinfo{title}{SportMamba: Adaptive Non-Linear Multi-Object Tracking with State Space Models for Team Sports}.
\newblock
\showeprint{arXiv:2506.03335}


\bibitem[Kono and Fujii(2024)]%
        {kono2024mathematical}
\bibfield{author}{\bibinfo{person}{Rikako Kono} {and} \bibinfo{person}{Keisuke Fujii}.} \bibinfo{year}{2024}\natexlab{}.
\newblock \showarticletitle{Mathematical models for off-ball scoring prediction in basketball}. In \bibinfo{booktitle}{\emph{International Workshop on Machine Learning and Data Mining for Sports Analytics}}. Springer.
\newblock


\bibitem[Koshkina and Elder(2024)]%
        {koshkina2024general}
\bibfield{author}{\bibinfo{person}{Maria Koshkina} {and} \bibinfo{person}{James~H Elder}.} \bibinfo{year}{2024}\natexlab{}.
\newblock \showarticletitle{A General Framework for Jersey Number Recognition in Sports Video}. In \bibinfo{booktitle}{\emph{Proceedings of the IEEE/CVF Conference on Computer Vision and Pattern Recognition}}. \bibinfo{pages}{3235--3244}.
\newblock


\bibitem[Koshkina et~al\mbox{.}(2021)]%
        {Koshkina2021-pn-hockey-classification}
\bibfield{author}{\bibinfo{person}{Maria Koshkina}, \bibinfo{person}{Hemanth Pidaparthy}, {and} \bibinfo{person}{James~H Elder}.} \bibinfo{year}{2021}\natexlab{}.
\newblock \showarticletitle{Contrastive learning for sports video: Unsupervised player classification}. In \bibinfo{booktitle}{\emph{Proceedings of the IEEE/CVF Conference on Computer Vision and Pattern Recognition}}. \bibinfo{pages}{4528--4536}.
\newblock


\bibitem[Kuang et~al\mbox{.}(2021)]%
        {kuang2021mmocr}
\bibfield{author}{\bibinfo{person}{Zhanghui Kuang}, \bibinfo{person}{Hongbin Sun}, \bibinfo{person}{Zhizhong Li}, \bibinfo{person}{Xiaoyu Yue}, \bibinfo{person}{Tsui~Hin Lin}, \bibinfo{person}{Jianyong Chen}, \bibinfo{person}{Huaqiang Wei}, \bibinfo{person}{Yiqin Zhu}, \bibinfo{person}{Tong Gao}, \bibinfo{person}{Wenwei Zhang}, {et~al\mbox{.}}} \bibinfo{year}{2021}\natexlab{}.
\newblock \showarticletitle{MMOCR: a comprehensive toolbox for text detection, recognition and understanding}. In \bibinfo{booktitle}{\emph{Proceedings of the 29th ACM International Conference on Multimedia}}. \bibinfo{pages}{3791--3794}.
\newblock


\bibitem[Kurach et~al\mbox{.}(2020)]%
        {kurach2020google}
\bibfield{author}{\bibinfo{person}{Karol Kurach}, \bibinfo{person}{Anton Raichuk}, \bibinfo{person}{Piotr Sta{\'n}czyk}, \bibinfo{person}{Micha{\l} Zaj{\k{a}}c}, \bibinfo{person}{Olivier Bachem}, \bibinfo{person}{Lasse Espeholt}, \bibinfo{person}{Carlos Riquelme}, \bibinfo{person}{Damien Vincent}, \bibinfo{person}{Marcin Michalski}, \bibinfo{person}{Olivier Bousquet}, {et~al\mbox{.}}} \bibinfo{year}{2020}\natexlab{}.
\newblock \showarticletitle{Google research football: A novel reinforcement learning environment}. In \bibinfo{booktitle}{\emph{Proceedings of the AAAI Conference on Artificial Intelligence}}, Vol.~\bibinfo{volume}{34}. \bibinfo{pages}{4501--4510}.
\newblock


\bibitem[Lin et~al\mbox{.}(2017)]%
        {lin2017focal}
\bibfield{author}{\bibinfo{person}{Tsung-Yi Lin}, \bibinfo{person}{Priya Goyal}, \bibinfo{person}{Ross Girshick}, \bibinfo{person}{Kaiming He}, {and} \bibinfo{person}{Piotr Doll{\'a}r}.} \bibinfo{year}{2017}\natexlab{}.
\newblock \showarticletitle{Focal loss for dense object detection}. In \bibinfo{booktitle}{\emph{Proceedings of the IEEE International Conference on Computer Vision}}. \bibinfo{pages}{2980--2988}.
\newblock


\bibitem[Lin et~al\mbox{.}(2015)]%
        {lin2015microsoft}
\bibfield{author}{\bibinfo{person}{Tsung-Yi Lin}, \bibinfo{person}{Michael Maire}, \bibinfo{person}{Serge Belongie}, \bibinfo{person}{Lubomir Bourdev}, \bibinfo{person}{Ross Girshick}, \bibinfo{person}{James Hays}, \bibinfo{person}{Pietro Perona}, \bibinfo{person}{Deva Ramanan}, \bibinfo{person}{C.~Lawrence Zitnick}, {and} \bibinfo{person}{Piotr Dollár}.} \bibinfo{year}{2015}\natexlab{}.
\newblock \bibinfo{title}{Microsoft COCO: Common Objects in Context}.
\newblock
\showeprint[arxiv]{1405.0312}~[cs.CV]


\bibitem[Lindstr{\"o}m et~al\mbox{.}(2020)]%
        {lindstrom2020predicting}
\bibfield{author}{\bibinfo{person}{Per Lindstr{\"o}m}, \bibinfo{person}{Ludwig Jacobsson}, \bibinfo{person}{Niklas Carlsson}, {and} \bibinfo{person}{Patrick Lambrix}.} \bibinfo{year}{2020}\natexlab{}.
\newblock \showarticletitle{Predicting player trajectories in shot situations in soccer}. In \bibinfo{booktitle}{\emph{Machine Learning and Data Mining for Sports Analytics: 7th International Workshop, MLSA 2020, Co-located with ECML/PKDD 2020, Ghent, Belgium, September 14--18, 2020, Proceedings 7}}. Springer, \bibinfo{pages}{62--75}.
\newblock


\bibitem[Liu et~al\mbox{.}(2023)]%
        {liu2023group}
\bibfield{author}{\bibinfo{person}{Huan Liu}, \bibinfo{person}{Qiang Chen}, \bibinfo{person}{Zichang Tan}, \bibinfo{person}{Jiang-Jiang Liu}, \bibinfo{person}{Jian Wang}, \bibinfo{person}{Xiangbo Su}, \bibinfo{person}{Xiaolong Li}, \bibinfo{person}{Kun Yao}, \bibinfo{person}{Junyu Han}, \bibinfo{person}{Errui Ding}, {et~al\mbox{.}}} \bibinfo{year}{2023}\natexlab{}.
\newblock \showarticletitle{Group pose: A simple baseline for end-to-end multi-person pose estimation}. In \bibinfo{booktitle}{\emph{Proceedings of the IEEE/CVF International Conference on Computer Vision}}. \bibinfo{pages}{15029--15038}.
\newblock


\bibitem[Lu et~al\mbox{.}(2017)]%
        {lu2017light}
\bibfield{author}{\bibinfo{person}{Keyu Lu}, \bibinfo{person}{Jianhui Chen}, \bibinfo{person}{James~J Little}, {and} \bibinfo{person}{Hangen He}.} \bibinfo{year}{2017}\natexlab{}.
\newblock \showarticletitle{Light cascaded convolutional neural networks for accurate player detection}. In \bibinfo{booktitle}{\emph{British Machine Vision Conference (BMVC)}}.
\newblock


\bibitem[Luiten et~al\mbox{.}(2021)]%
        {luiten2021hota}
\bibfield{author}{\bibinfo{person}{Jonathon Luiten}, \bibinfo{person}{Aljosa Osep}, \bibinfo{person}{Patrick Dendorfer}, \bibinfo{person}{Philip Torr}, \bibinfo{person}{Andreas Geiger}, \bibinfo{person}{Laura Leal-Taix{\'e}}, {and} \bibinfo{person}{Bastian Leibe}.} \bibinfo{year}{2021}\natexlab{}.
\newblock \showarticletitle{Hota: A higher order metric for evaluating multi-object tracking}.
\newblock \bibinfo{journal}{\emph{International journal of computer vision}}  \bibinfo{volume}{129} (\bibinfo{year}{2021}), \bibinfo{pages}{548--578}.
\newblock


\bibitem[Mackowiak et~al\mbox{.}(2010)]%
        {mackowiak2010football}
\bibfield{author}{\bibinfo{person}{Slawomir Mackowiak}, \bibinfo{person}{Jacek Konieczny}, \bibinfo{person}{Maciej Kurc}, {and} \bibinfo{person}{Przemysław Maćkowiak}.} \bibinfo{year}{2010}\natexlab{}.
\newblock \showarticletitle{Football player detection in video broadcast}.
\newblock \bibinfo{journal}{\emph{Computer Vision and Graphics, Lecture Notes in Computer Science}}  \bibinfo{volume}{6375} (\bibinfo{year}{2010}), \bibinfo{pages}{118--125}.
\newblock


\bibitem[Maglo et~al\mbox{.}(2022)]%
        {maglo2022efficient}
\bibfield{author}{\bibinfo{person}{Adrien Maglo}, \bibinfo{person}{Astrid Orcesi}, {and} \bibinfo{person}{Quoc-Cuong Pham}.} \bibinfo{year}{2022}\natexlab{}.
\newblock \showarticletitle{Efficient tracking of team sport players with few game-specific annotations}. In \bibinfo{booktitle}{\emph{Proceedings of the IEEE/CVF Conference on Computer Vision and Pattern Recognition}}. \bibinfo{pages}{3461--3471}.
\newblock


\bibitem[Majeed et~al\mbox{.}(2024)]%
        {majeed2024mv}
\bibfield{author}{\bibinfo{person}{Fahad Majeed}, \bibinfo{person}{Nauman~Ullah Gilal}, \bibinfo{person}{Khaled Al-Thelaya}, \bibinfo{person}{Yin Yang}, \bibinfo{person}{Marco Agus}, {and} \bibinfo{person}{Jens Schneider}.} \bibinfo{year}{2024}\natexlab{}.
\newblock \showarticletitle{MV-Soccer: Motion-Vector Augmented Instance Segmentation for Soccer Player Tracking}. In \bibinfo{booktitle}{\emph{Proceedings of the IEEE/CVF Conference on Computer Vision and Pattern Recognition}}. \bibinfo{pages}{3245--3255}.
\newblock


\bibitem[Mansourian et~al\mbox{.}(2023)]%
        {mansourian2023multi}
\bibfield{author}{\bibinfo{person}{Amir~M Mansourian}, \bibinfo{person}{Vladimir Somers}, \bibinfo{person}{Christophe De~Vleeschouwer}, {and} \bibinfo{person}{Shohreh Kasaei}.} \bibinfo{year}{2023}\natexlab{}.
\newblock \showarticletitle{Multi-task learning for joint re-identification, team affiliation, and role classification for sports visual tracking}. In \bibinfo{booktitle}{\emph{Proceedings of the 6th International Workshop on Multimedia Content Analysis in Sports}}. \bibinfo{pages}{103--112}.
\newblock


\bibitem[Nakahara et~al\mbox{.}(2023)]%
        {nakahara2023action}
\bibfield{author}{\bibinfo{person}{Hiroshi Nakahara}, \bibinfo{person}{Kazushi Tsutsui}, \bibinfo{person}{Kazuya Takeda}, {and} \bibinfo{person}{Keisuke Fujii}.} \bibinfo{year}{2023}\natexlab{}.
\newblock \showarticletitle{Action valuation of on-and off-ball soccer players based on multi-agent deep reinforcement learning}.
\newblock \bibinfo{journal}{\emph{IEEE Access}}  \bibinfo{volume}{11} (\bibinfo{year}{2023}), \bibinfo{pages}{131237--131244}.
\newblock


\bibitem[Penate-Sanchez et~al\mbox{.}(2020)]%
        {penate2020tgc20reid}
\bibfield{author}{\bibinfo{person}{Adrian Penate-Sanchez}, \bibinfo{person}{David Freire-Obregon}, \bibinfo{person}{Adrian Lorenzo-Melian}, \bibinfo{person}{Javier Lorenzo-Navarro}, {and} \bibinfo{person}{Modesto Castrillon-Santana}.} \bibinfo{year}{2020}\natexlab{}.
\newblock \showarticletitle{{TGC20R}e{I}d: A dataset for sport event re-identification in the wild}.
\newblock \bibinfo{journal}{\emph{Pattern Recognition Letters}}  \bibinfo{volume}{138} (\bibinfo{year}{2020}), \bibinfo{pages}{355--361}.
\newblock


\bibitem[Pettersen et~al\mbox{.}(2014)]%
        {pettersen2014soccer}
\bibfield{author}{\bibinfo{person}{Svein~Arne Pettersen}, \bibinfo{person}{Dag Johansen}, \bibinfo{person}{H{\aa}vard Johansen}, \bibinfo{person}{Vegard Berg-Johansen}, \bibinfo{person}{Vamsidhar~Reddy Gaddam}, \bibinfo{person}{Asgeir Mortensen}, \bibinfo{person}{Ragnar Langseth}, \bibinfo{person}{Carsten Griwodz}, \bibinfo{person}{H{\aa}kon~Kvale Stensland}, {and} \bibinfo{person}{P{\aa}l Halvorsen}.} \bibinfo{year}{2014}\natexlab{}.
\newblock \showarticletitle{Soccer video and player position dataset}. In \bibinfo{booktitle}{\emph{Proceedings of the 5th ACM Multimedia Systems Conference}}. \bibinfo{pages}{18--23}.
\newblock


\bibitem[Pishchulin et~al\mbox{.}(2016)]%
        {pishchulin2016deepcut}
\bibfield{author}{\bibinfo{person}{Leonid Pishchulin}, \bibinfo{person}{Eldar Insafutdinov}, \bibinfo{person}{Siyu Tang}, \bibinfo{person}{Bjoern Andres}, \bibinfo{person}{Mykhaylo Andriluka}, \bibinfo{person}{Peter~V Gehler}, {and} \bibinfo{person}{Bernt Schiele}.} \bibinfo{year}{2016}\natexlab{}.
\newblock \showarticletitle{Deepcut: Joint subset partition and labeling for multi person pose estimation}. In \bibinfo{booktitle}{\emph{Proceedings of the IEEE Conference on Computer Vision and Pattern Recognition}}. \bibinfo{pages}{4929--4937}.
\newblock


\bibitem[Qin et~al\mbox{.}(2025)]%
        {qin2025soccersynth}
\bibfield{author}{\bibinfo{person}{Haobin Qin}, \bibinfo{person}{Calvin Yeung}, \bibinfo{person}{Rikuhei Umemoto}, {and} \bibinfo{person}{Keisuke Fujii}.} \bibinfo{year}{2025}\natexlab{}.
\newblock \showarticletitle{SoccerSynth-Detection: A Synthetic Dataset for Soccer Player Detection}.
\newblock \bibinfo{journal}{\emph{arXiv preprint arXiv:2501.09281}} (\bibinfo{year}{2025}).
\newblock


\bibitem[Redmon et~al\mbox{.}(2016)]%
        {redmon2016you}
\bibfield{author}{\bibinfo{person}{Joseph Redmon}, \bibinfo{person}{Santosh Divvala}, \bibinfo{person}{Ross Girshick}, {and} \bibinfo{person}{Ali Farhadi}.} \bibinfo{year}{2016}\natexlab{}.
\newblock \showarticletitle{You only look once: Unified, real-time object detection}. In \bibinfo{booktitle}{\emph{Proceedings of the IEEE Conference on Computer Vision and Pattern Recognition}}. \bibinfo{pages}{779--788}.
\newblock


\bibitem[Scott et~al\mbox{.}(2024)]%
        {scott2024teamtrack}
\bibfield{author}{\bibinfo{person}{Atom Scott}, \bibinfo{person}{Ikuma Uchida}, \bibinfo{person}{Ning Ding}, \bibinfo{person}{Rikuhei Umemoto}, \bibinfo{person}{Rory Bunker}, \bibinfo{person}{Ren Kobayashi}, \bibinfo{person}{Takeshi Koyama}, \bibinfo{person}{Masaki Onishi}, \bibinfo{person}{Yoshinari Kameda}, {and} \bibinfo{person}{Keisuke Fujii}.} \bibinfo{year}{2024}\natexlab{}.
\newblock \showarticletitle{TeamTrack: A Dataset for Multi-Sport Multi-Object Tracking in Full-pitch Videos}. In \bibinfo{booktitle}{\emph{Proceedings of the IEEE/CVF Conference on Computer Vision and Pattern Recognition}}. \bibinfo{pages}{3357--3366}.
\newblock


\bibitem[Scott et~al\mbox{.}(2022)]%
        {scott2022soccertrack}
\bibfield{author}{\bibinfo{person}{Atom Scott}, \bibinfo{person}{Ikuma Uchida}, \bibinfo{person}{Masaki Onishi}, \bibinfo{person}{Yoshinari Kameda}, \bibinfo{person}{Kazuhiro Fukui}, {and} \bibinfo{person}{Keisuke Fujii}.} \bibinfo{year}{2022}\natexlab{}.
\newblock \showarticletitle{SoccerTrack: A Dataset and Tracking Algorithm for Soccer With Fish-Eye and Drone Videos}. In \bibinfo{booktitle}{\emph{8th International Workshop on Computer Vision in Sports (CVsports) at IEEE/CVF Conference on Computer Vision and Pattern Recognition (CVPR' 22)}}. \bibinfo{pages}{3569--3579}.
\newblock


\bibitem[Shi et~al\mbox{.}(2022)]%
        {shi2022end}
\bibfield{author}{\bibinfo{person}{Dahu Shi}, \bibinfo{person}{Xing Wei}, \bibinfo{person}{Liangqi Li}, \bibinfo{person}{Ye Ren}, {and} \bibinfo{person}{Wenming Tan}.} \bibinfo{year}{2022}\natexlab{}.
\newblock \showarticletitle{End-to-end multi-person pose estimation with transformers}. In \bibinfo{booktitle}{\emph{Proceedings of the IEEE/CVF Conference on Computer Vision and Pattern Recognition}}. \bibinfo{pages}{11069--11078}.
\newblock


\bibitem[Somers et~al\mbox{.}(2024a)]%
        {somers2024keypoint}
\bibfield{author}{\bibinfo{person}{Vladimir Somers}, \bibinfo{person}{Alexandre Alahi}, {and} \bibinfo{person}{Christophe~De Vleeschouwer}.} \bibinfo{year}{2024}\natexlab{a}.
\newblock \showarticletitle{Keypoint promptable re-identification}. In \bibinfo{booktitle}{\emph{European Conference on Computer Vision}}. Springer, \bibinfo{pages}{216--233}.
\newblock


\bibitem[Somers et~al\mbox{.}(2023)]%
        {somers2023body}
\bibfield{author}{\bibinfo{person}{Vladimir Somers}, \bibinfo{person}{Christophe De~Vleeschouwer}, {and} \bibinfo{person}{Alexandre Alahi}.} \bibinfo{year}{2023}\natexlab{}.
\newblock \showarticletitle{Body part-based representation learning for occluded person re-identification}. In \bibinfo{booktitle}{\emph{Proceedings of the IEEE/CVF winter conference on applications of computer vision}}. \bibinfo{pages}{1613--1623}.
\newblock


\bibitem[Somers et~al\mbox{.}(2024b)]%
        {somers2024soccernet}
\bibfield{author}{\bibinfo{person}{Vladimir Somers}, \bibinfo{person}{Victor Joos}, \bibinfo{person}{Anthony Cioppa}, \bibinfo{person}{Silvio Giancola}, \bibinfo{person}{Seyed~Abolfazl Ghasemzadeh}, \bibinfo{person}{Floriane Magera}, \bibinfo{person}{Baptiste Standaert}, \bibinfo{person}{Amir~M Mansourian}, \bibinfo{person}{Xin Zhou}, \bibinfo{person}{Shohreh Kasaei}, {et~al\mbox{.}}} \bibinfo{year}{2024}\natexlab{b}.
\newblock \showarticletitle{SoccerNet game state reconstruction: End-to-end athlete tracking and identification on a minimap}. In \bibinfo{booktitle}{\emph{Proceedings of the IEEE/CVF Conference on Computer Vision and Pattern Recognition}}. \bibinfo{pages}{3293--3305}.
\newblock


\bibitem[Somers et~al\mbox{.}(2025)]%
        {somers2025CAMELTrack}
\bibfield{author}{\bibinfo{person}{Vladimir Somers}, \bibinfo{person}{Baptiste Standaert}, \bibinfo{person}{Victor Joos}, \bibinfo{person}{Alexandre Alahi}, {and} \bibinfo{person}{Christophe~De Vleeschouwer}.} \bibinfo{year}{2025}\natexlab{}.
\newblock \bibinfo{title}{CAMELTrack: Context-Aware Multi-cue ExpLoitation for Online Multi-Object Tracking}.
\newblock
\showeprint[arxiv]{2505.01257}~[cs.CV]
\urldef\tempurl%
\url{https://arxiv.org/abs/2505.01257}
\showURL{%
\tempurl}


\bibitem[Spearman(2018)]%
        {Spearman18}
\bibfield{author}{\bibinfo{person}{William Spearman}.} \bibinfo{year}{2018}\natexlab{}.
\newblock \showarticletitle{Beyond expected goals}. In \bibinfo{booktitle}{\emph{Proceedings of the 12th MIT Sloan Sports Analytics Conference}}. \bibinfo{pages}{1--17}.
\newblock


\bibitem[Stanczyk et~al\mbox{.}(2025)]%
        {Stanczyk2025NoTrainYetGain}
\bibfield{author}{\bibinfo{person}{Tomasz Stanczyk}, \bibinfo{person}{Seongro Yoon}, {and} \bibinfo{person}{Francois Bremond}.} \bibinfo{year}{2025}\natexlab{}.
\newblock \bibinfo{title}{No Train Yet Gain: Towards Generic Multi-Object Tracking in Sports and Beyond}.
\newblock
\showeprint{arXiv:2506.01373}


\bibitem[Sun et~al\mbox{.}(2024)]%
        {sun2024gta}
\bibfield{author}{\bibinfo{person}{Jiacheng Sun}, \bibinfo{person}{Hsiang-Wei Huang}, \bibinfo{person}{Cheng-Yen Yang}, \bibinfo{person}{Zhongyu Jiang}, {and} \bibinfo{person}{Jenq-Neng Hwang}.} \bibinfo{year}{2024}\natexlab{}.
\newblock \showarticletitle{GTA: Global Tracklet Association for Multi-Object Tracking in Sports}. In \bibinfo{booktitle}{\emph{Proceedings of the Asian Conference on Computer Vision}}. \bibinfo{pages}{421--434}.
\newblock


\bibitem[Sun et~al\mbox{.}(2019)]%
        {Sun2019Deep}
\bibfield{author}{\bibinfo{person}{Ke Sun}, \bibinfo{person}{Bin Xiao}, \bibinfo{person}{Dong Liu}, {and} \bibinfo{person}{Jingdong Wang}.} \bibinfo{year}{2019}\natexlab{}.
\newblock \showarticletitle{Deep high-resolution representation learning for human pose estimation}. In \bibinfo{booktitle}{\emph{Proceedings of the IEEE/CVF conference on computer vision and pattern recognition}}. \bibinfo{pages}{5693--5703}.
\newblock


\bibitem[Sun et~al\mbox{.}(2022)]%
        {sun2022dancetrack}
\bibfield{author}{\bibinfo{person}{Peize Sun}, \bibinfo{person}{Jinkun Cao}, \bibinfo{person}{Yi Jiang}, \bibinfo{person}{Zehuan Yuan}, \bibinfo{person}{Song Bai}, \bibinfo{person}{Kris Kitani}, {and} \bibinfo{person}{Ping Luo}.} \bibinfo{year}{2022}\natexlab{}.
\newblock \showarticletitle{Dancetrack: Multi-object tracking in uniform appearance and diverse motion}. In \bibinfo{booktitle}{\emph{Proceedings of the IEEE/CVF Conference on Computer Vision and Pattern Recognition}}. \bibinfo{pages}{20993--21002}.
\newblock


\bibitem[Suzuki et~al\mbox{.}(2024)]%
        {suzuki2024runner}
\bibfield{author}{\bibinfo{person}{Tomohiro Suzuki}, \bibinfo{person}{Kazushi Tsutsui}, \bibinfo{person}{Kazuya Takeda}, {and} \bibinfo{person}{Keisuke Fujii}.} \bibinfo{year}{2024}\natexlab{}.
\newblock \showarticletitle{Runner re-identification from single-view running video in the open-world setting}.
\newblock \bibinfo{journal}{\emph{Multimedia Tools and Applications}} (\bibinfo{year}{2024}), \bibinfo{pages}{1--17}.
\newblock


\bibitem[Theiner and Ewerth(2023)]%
        {theiner2023tvcalib}
\bibfield{author}{\bibinfo{person}{Jonas Theiner} {and} \bibinfo{person}{Ralph Ewerth}.} \bibinfo{year}{2023}\natexlab{}.
\newblock \showarticletitle{Tvcalib: Camera calibration for sports field registration in soccer}. In \bibinfo{booktitle}{\emph{Proceedings of the IEEE/CVF Winter Conference on Applications of Computer Vision}}. \bibinfo{pages}{1166--1175}.
\newblock


\bibitem[Toda et~al\mbox{.}(2022)]%
        {toda2022evaluation}
\bibfield{author}{\bibinfo{person}{Kosuke Toda}, \bibinfo{person}{Masakiyo Teranishi}, \bibinfo{person}{Keisuke Kushiro}, {and} \bibinfo{person}{Keisuke Fujii}.} \bibinfo{year}{2022}\natexlab{}.
\newblock \showarticletitle{Evaluation of soccer team defense based on prediction models of ball recovery and being attacked: A pilot study}.
\newblock \bibinfo{journal}{\emph{PLoS One}} \bibinfo{volume}{17}, \bibinfo{number}{1} (\bibinfo{year}{2022}), \bibinfo{pages}{e0263051}.
\newblock


\bibitem[Toshev and Szegedy(2014)]%
        {Toshev2014DeepPose}
\bibfield{author}{\bibinfo{person}{Alexander Toshev} {and} \bibinfo{person}{Christian Szegedy}.} \bibinfo{year}{2014}\natexlab{}.
\newblock \showarticletitle{Deeppose: Human pose estimation via deep neural networks}. In \bibinfo{booktitle}{\emph{Proceedings of the IEEE Conference on Computer Vision and Pattern Recognition}}. \bibinfo{pages}{1653--1660}.
\newblock


\bibitem[Van~Zandycke et~al\mbox{.}(2022)]%
        {van2022deepsportradarv1}
\bibfield{author}{\bibinfo{person}{Gabriel Van~Zandycke}, \bibinfo{person}{Vladimir Somers}, \bibinfo{person}{Maxime Istasse}, \bibinfo{person}{Carlo~Del Don}, {and} \bibinfo{person}{Davide Zambrano}.} \bibinfo{year}{2022}\natexlab{}.
\newblock \showarticletitle{DeepSportradar-v1: Computer vision dataset for sports understanding with high quality annotations}. In \bibinfo{booktitle}{\emph{Proceedings of the 5th International ACM Workshop on Multimedia Content Analysis in Sports}}. \bibinfo{pages}{1--8}.
\newblock


\bibitem[Vats et~al\mbox{.}(2022)]%
        {Vats2022-lq-hockey-tfidentification}
\bibfield{author}{\bibinfo{person}{Kanav Vats}, \bibinfo{person}{William McNally}, \bibinfo{person}{Pascale Walters}, \bibinfo{person}{David~A Clausi}, {and} \bibinfo{person}{John~S Zelek}.} \bibinfo{year}{2022}\natexlab{}.
\newblock \showarticletitle{Ice hockey player identification via transformers and weakly supervised learning}. In \bibinfo{booktitle}{\emph{Proceedings of the IEEE/CVF Conference on Computer Vision and Pattern Recognition}}. \bibinfo{pages}{3451--3460}.
\newblock


\bibitem[Vats et~al\mbox{.}(2023)]%
        {vats2023player}
\bibfield{author}{\bibinfo{person}{Kanav Vats}, \bibinfo{person}{Pascale Walters}, \bibinfo{person}{Mehrnaz Fani}, \bibinfo{person}{David~A Clausi}, {and} \bibinfo{person}{John~S Zelek}.} \bibinfo{year}{2023}\natexlab{}.
\newblock \showarticletitle{Player tracking and identification in ice hockey}.
\newblock \bibinfo{journal}{\emph{Expert Systems with Applications}}  \bibinfo{volume}{213} (\bibinfo{year}{2023}), \bibinfo{pages}{119250}.
\newblock


\bibitem[Wang et~al\mbox{.}(2011)]%
        {wang2011learning}
\bibfield{author}{\bibinfo{person}{Yang Wang}, \bibinfo{person}{Duan Tran}, {and} \bibinfo{person}{Zicheng Liao}.} \bibinfo{year}{2011}\natexlab{}.
\newblock \showarticletitle{Learning hierarchical poselets for human parsing}. In \bibinfo{booktitle}{\emph{CVPR 2011}}. IEEE, \bibinfo{pages}{1705--1712}.
\newblock


\bibitem[Wojke et~al\mbox{.}(2017)]%
        {wojke2017simple}
\bibfield{author}{\bibinfo{person}{Nicolai Wojke}, \bibinfo{person}{Alex Bewley}, {and} \bibinfo{person}{Dietrich Paulus}.} \bibinfo{year}{2017}\natexlab{}.
\newblock \showarticletitle{Simple online and realtime tracking with a deep association metric}. In \bibinfo{booktitle}{\emph{2017 IEEE International Conference on Image Processing (ICIP)}}. IEEE, \bibinfo{pages}{3645--3649}.
\newblock


\bibitem[Xiong et~al\mbox{.}(2022)]%
        {xiong2022swin}
\bibfield{author}{\bibinfo{person}{Zinan Xiong}, \bibinfo{person}{Chenxi Wang}, \bibinfo{person}{Ying Li}, \bibinfo{person}{Yan Luo}, {and} \bibinfo{person}{Yu Cao}.} \bibinfo{year}{2022}\natexlab{}.
\newblock \showarticletitle{Swin-pose: Swin transformer based human pose estimation}. In \bibinfo{booktitle}{\emph{2022 IEEE 5th International Conference on Multimedia Information Processing and Retrieval (MIPR)}}. IEEE, \bibinfo{pages}{228--233}.
\newblock


\bibitem[Xu et~al\mbox{.}(2024)]%
        {xu2024finesports}
\bibfield{author}{\bibinfo{person}{Jinglin Xu}, \bibinfo{person}{Guohao Zhao}, \bibinfo{person}{Sibo Yin}, \bibinfo{person}{Wenhao Zhou}, {and} \bibinfo{person}{Yuxin Peng}.} \bibinfo{year}{2024}\natexlab{}.
\newblock \showarticletitle{FineSports: A Multi-person Hierarchical Sports Video Dataset for Fine-grained Action Understanding}. In \bibinfo{booktitle}{\emph{Proceedings of the IEEE/CVF Conference on Computer Vision and Pattern Recognition}}. \bibinfo{pages}{21773--21782}.
\newblock


\bibitem[Xu et~al\mbox{.}(2022)]%
        {xu2022vitpose}
\bibfield{author}{\bibinfo{person}{Yufei Xu}, \bibinfo{person}{Jing Zhang}, \bibinfo{person}{Qiming Zhang}, {and} \bibinfo{person}{Dacheng Tao}.} \bibinfo{year}{2022}\natexlab{}.
\newblock \showarticletitle{Vitpose: Simple vision transformer baselines for human pose estimation}.
\newblock \bibinfo{journal}{\emph{Advances in Neural Information Processing Systems}}  \bibinfo{volume}{35} (\bibinfo{year}{2022}), \bibinfo{pages}{38571--38584}.
\newblock


\bibitem[Yeung and Fujii(2024)]%
        {yeung2024strategic}
\bibfield{author}{\bibinfo{person}{Calvin Yeung} {and} \bibinfo{person}{Keisuke Fujii}.} \bibinfo{year}{2024}\natexlab{}.
\newblock \showarticletitle{A strategic framework for optimal decisions in football 1-vs-1 shot-taking situations: An integrated approach of machine learning, theory-based modeling, and game theory}.
\newblock \bibinfo{journal}{\emph{Complex \& Intelligent Systems}} (\bibinfo{year}{2024}), \bibinfo{pages}{1--20}.
\newblock


\bibitem[Yeung et~al\mbox{.}(2024)]%
        {yeung2024autosoccerpose}
\bibfield{author}{\bibinfo{person}{Calvin Yeung}, \bibinfo{person}{Kenjiro Ide}, {and} \bibinfo{person}{Keisuke Fujii}.} \bibinfo{year}{2024}\natexlab{}.
\newblock \showarticletitle{AutoSoccerPose: Automated 3D posture Analysis of Soccer Shot Movements}. In \bibinfo{booktitle}{\emph{Proceedings of the IEEE/CVF Conference on Computer Vision and Pattern Recognition}}. \bibinfo{pages}{3214--3224}.
\newblock


\bibitem[Yeung et~al\mbox{.}(2025a)]%
        {yeung2025openstarlab}
\bibfield{author}{\bibinfo{person}{Calvin Yeung}, \bibinfo{person}{Kenjiro Ide}, \bibinfo{person}{Taiga Someya}, {and} \bibinfo{person}{Keisuke Fujii}.} \bibinfo{year}{2025}\natexlab{a}.
\newblock \showarticletitle{OpenSTARLab: Open Approach for Spatio-Temporal Agent Data Analysis in Soccer}.
\newblock \bibinfo{journal}{\emph{arXiv preprint arXiv:2502.02785}} (\bibinfo{year}{2025}).
\newblock


\bibitem[Yeung et~al\mbox{.}(2025b)]%
        {yeung2025transformer}
\bibfield{author}{\bibinfo{person}{Calvin Yeung}, \bibinfo{person}{Tony Sit}, {and} \bibinfo{person}{Keisuke Fujii}.} \bibinfo{year}{2025}\natexlab{b}.
\newblock \showarticletitle{Transformer-based neural marked spatio temporal point process model for analyzing football match events}.
\newblock \bibinfo{journal}{\emph{Applied Intelligence}} \bibinfo{volume}{55}, \bibinfo{number}{5} (\bibinfo{year}{2025}), \bibinfo{pages}{1--17}.
\newblock


\bibitem[Yin et~al\mbox{.}(2024)]%
        {yin2024enhanced}
\bibfield{author}{\bibinfo{person}{Li Yin}, \bibinfo{person}{Calvin Yeung}, \bibinfo{person}{Qingrui Hu}, \bibinfo{person}{Jun Ichikawa}, \bibinfo{person}{Hirotsugu Azechi}, \bibinfo{person}{Susumu Takahashi}, {and} \bibinfo{person}{Keisuke Fujii}.} \bibinfo{year}{2024}\natexlab{}.
\newblock \showarticletitle{Enhanced Multi-Object Tracking Using Pose-based Virtual Markers in 3x3 Basketball}.
\newblock \bibinfo{journal}{\emph{arXiv preprint arXiv:2412.06258}} (\bibinfo{year}{2024}).
\newblock


\bibitem[Zhang et~al\mbox{.}(2022)]%
        {zhang2022bytetrack}
\bibfield{author}{\bibinfo{person}{Yifu Zhang}, \bibinfo{person}{Peize Sun}, \bibinfo{person}{Yi Jiang}, \bibinfo{person}{Dongdong Yu}, \bibinfo{person}{Fucheng Weng}, \bibinfo{person}{Zehuan Yuan}, \bibinfo{person}{Ping Luo}, \bibinfo{person}{Wenyu Liu}, {and} \bibinfo{person}{Xinggang Wang}.} \bibinfo{year}{2022}\natexlab{}.
\newblock \showarticletitle{Bytetrack: Multi-object tracking by associating every detection box}. In \bibinfo{booktitle}{\emph{European Conference on Computer Vision}}. Springer, \bibinfo{pages}{1--21}.
\newblock


\bibitem[Zhao et~al\mbox{.}(2020)]%
        {yingnan2020MHPTD}
\bibfield{author}{\bibinfo{person}{Yingnan Zhao}, \bibinfo{person}{Zihui Li}, {and} \bibinfo{person}{Kua Chen}.} \bibinfo{year}{2020}\natexlab{}.
\newblock \showarticletitle{A Method for Tracking Hockey Players by Exploiting Multiple Detections and Omni-Scale Appearance Features}.
\newblock \bibinfo{journal}{\emph{Project Report}} (\bibinfo{year}{2020}).
\newblock


\end{thebibliography}

\clearpage
\appendix
\renewcommand{\thetable}{\Alph{section}.\arabic{figure}}
\renewcommand{\thefigure}{\Alph{section}.\arabic{table}}
\section{MOT experiment on Drone dataset}
\label{app:MOTdrone}
We describe a simple baseline for player tracking in Drone dataset. 
We implement two conventional baseline MOT algorithms: ByteTrack \cite{zhang2022bytetrack} and BoT-SORT \cite{aharon2022bot}. 

For performance evaluation, we use HOTA \cite{luiten2021hota} metric due to its comprehensive ability to measure both detection and association performance. Additionally, we incorporate supplementary metrics, including DetA, AssA, and ID switches (IDs), to provide a more granular analysis of tracking performance.  

Table \ref{tab:byte_bot_drone} shows the results with the default parameters of the two methods. The results reveal that the BoT-SORT method demonstrates improvement in HOTA scores compared to ByteTrack, along with a reduction in ID switches. However, there remains considerable room for further performance enhancement due to many occlusions.

\begin{table}[!b]
\centering
\scalebox{0.69}{
\begin{tabular}{cccccc}
\hline
\textbf{Method}                     & \textbf{Length [s]} & \textbf{HOTA(\%)↑} & \textbf{DetA(\%)↑}        & \textbf{AssA(\%)↑}      & \textbf{IDs↓} \\
\hline
ByteTrack \cite{zhang2022bytetrack} & 39.55 $\pm$ 36.30    & 42.92 $\pm$ 11.96   & 37.02 $\pm$ 8.86  & 51.77 $\pm$ 22.21 & 15.06 $\pm$ 14.22 \\
BoT-SORT \cite{aharon2022bot}  & 39.55 $\pm$ 36.30    & 49.98 $\pm$ 12.24   & 42.22 $\pm$ 9.63 & 61.82 $\pm$ 22.46  & 11.81 $\pm$ 13.19 \\
\hline
\end{tabular}
}
\caption{ByteTrack and BoT-SORT results on Drone dataset.}
\label{tab:byte_bot_drone}
\vspace{-15pt}
\end{table}

\end{document}